\documentclass[pdflatex,sn-basic]{sn-jnl}

\usepackage{graphicx}%
\usepackage{tabularx}
\usepackage{multirow}%
\usepackage{amsmath,amssymb,amsfonts}%
\usepackage{amsthm}%
\usepackage{mathrsfs}%
\usepackage[title]{appendix}%
\usepackage{xcolor}%
\usepackage{textcomp}%
\usepackage{manyfoot}%
\usepackage{booktabs}%
\usepackage{algorithm}%
\usepackage{algorithmicx}%
\usepackage{algpseudocode}%
\usepackage{listings}%
\usepackage{adjustbox}
\usepackage{pifont}
\usepackage{makecell}
\usepackage{multirow} 
\usepackage{graphicx}
\usepackage{geometry}
\usepackage{subcaption}
\usepackage{xurl}
\usepackage{hyperref}
\usepackage{array}
\usepackage{color}
\usepackage{xspace}
\usepackage{colortbl}
\definecolor{meddr}{HTML}{C73A32}
\usepackage{tcolorbox}
\usepackage{titlesec}

\newcommand\ours{MedDr}

\definecolor{green1}{RGB}{84, 130, 53}
\definecolor{red1}{RGB}{192, 0, 0}
\definecolor{orange1}{RGB}{237, 125, 49}

\begin{document}

\title[Article Title]{GSCo: Towards Generalizable AI in Medicine via Generalist-Specialist Collaboration}

\author[1]{\fnm{Sunan} \sur{He}}
\equalcont{These authors contributed equally to this work.}
\author[1]{\fnm{Yuxiang} \sur{Nie}}
\equalcont{These authors contributed equally to this work.}
\author[1]{\fnm{Hongmei} \sur{Wang}}
\author[1]{\fnm{Shu} \sur{Yang}}
\author[1]{\fnm{Yihui} \sur{Wang}}
\author[1]{\fnm{Zhiyuan} \sur{Cai}}
\author[1]{\fnm{Zhixuan} \sur{Chen}}
\author[1]{\fnm{Yingxue} \sur{Xu}}
\author[2]{\fnm{Luyang} \sur{Luo}}
\author[3]{\fnm{Huiling} \sur{Xiang}}
\author[3]{\fnm{Xi} \sur{Lin}}
\author[4]{\fnm{Mingxiang} \sur{Wu}}
\author[5]{\fnm{Yifan} \sur{Peng}}
\author[6]{\fnm{George} \sur{Shih}}
\author[7]{\fnm{Ziyang} \sur{Xu}}
\author[8]{\fnm{Xian} \sur{Wu}}
\author[9]{\fnm{Qiong} \sur{Wang}}
\author[10, 11]{\fnm{Ronald Cheong Kin} \sur{Chan}}
\author[12]{\fnm{Varut} \sur{Vardhanabhuti}}
\author[13]{\fnm{Winnie Chiu Wing} \sur{Chu}}
\author[14]{\fnm{Yefeng} \sur{Zheng}}
\author[2]{\fnm{Pranav} \sur{Rajpurkar}}
\author[15]{\fnm{Kang} \sur{Zhang}}
\author*[1,16,17,18,19]{\fnm{Hao} \sur{Chen}}\email{jhc@cse.ust.hk}

\affil[1]{\orgdiv{Department of Computer Science and Engineering}, \orgname{The Hong Kong University of Science and Technology}, \orgaddress{\city{Hong Kong}, \country{China}}}

\affil[2]{\orgdiv{Department of Biomedical Informatics}, \orgname{Harvard University}, \orgaddress{\city{Boston}, \country{USA}}}

\affil[3]{\orgdiv{Department of Ultrasound}, \orgname{Sun Yat-sen University Cancer Center}, \orgaddress{\city{Guangzhou}, \country{China}}}

\affil[4]{\orgdiv{Department of Radiology}, \orgname{Shenzhen People’s Hospital}, \orgaddress{\city{Shenzhen}, \country{China}}}

\affil[5]{\orgdiv{Population Health Sciences}, \orgname{Weill Cornell Medicine}, \orgaddress{\city{New York}, \country{USA}}}

\affil[6]{\orgdiv{Department of Radiology}, \orgname{Weill Cornell Medicine}, \orgaddress{\city{New York}, \country{USA}}}

\affil[7]{\orgdiv{Perelman Department of Dermatology}, \orgname{New York Langone Health}, \orgaddress{\city{New York}, \country{USA}}}

\affil[8]{\orgdiv{Jarvis Research Center}, \orgname{Tencent YouTu Lab}, \orgaddress{\city{Shenzhen}, \country{China}}}

\affil[9]{\orgdiv{Shenzhen Institute of Advanced
Technology}, \orgname{Chinese Academy of Sciences}, \orgaddress{\city{Shenzhen}, \country{China}}}

\affil[10]{\orgdiv{Department of Anatomical and Cellular Pathology}, \orgname{The Chinese University of Hong Kong}, \orgaddress{\city{Hong Kong}, \country{China}}}

\affil[11]{\orgdiv{State Key Laboratory of Translational Oncology}, \orgname{The Chinese University of Hong Kong}, \orgaddress{\city{Hong Kong}, \country{China}}}

\affil[12]{\orgdiv{Department of Diagnostic Radiology}, \orgname{The University of Hong Kong}, \orgaddress{\city{Hong Kong}, \country{China}}}

\affil[13]{\orgdiv{Department of Imaging and Interventional Radiology}, \orgname{The Chinese University of Hong Kong}, \orgaddress{\city{Hong Kong}, \country{China}}}

\affil[14]{\orgdiv{Medical Artificial Intelligence Laboratory}, \orgname{Westlake University}, \orgaddress{\city{Hangzhou}, \country{China}}}

\affil[15]{\orgdiv{Faculty of Medicine}, \orgname{The Macau University of Science and Technology}, \orgaddress{\city{Macao}, \country{China}}}

\affil[16]{\orgdiv{Department of Chemical and Biological Engineering}, \orgname{The Hong Kong University of Science and Technology}, \orgaddress{\city{Hong Kong}, \country{China}}}

\affil[17]{\orgdiv{Division of Life Science}, \orgname{The Hong Kong University of Science and Technology}, \orgaddress{\city{Hong Kong}, \country{China}}}

\affil[18]{\orgdiv{State Key Laboratory of Molecular Neuroscience}, \orgname{The Hong Kong University of Science and Technology}, \orgaddress{\city{Hong Kong}, \country{China}}}

\affil[19]{\orgdiv{Shenzhen-Hong Kong Collaborative Innovation Research Institute}, \orgname{The Hong Kong University of Science and Technology}, \orgaddress{\city{Shenzhen}, \country{China}}}


\newtcolorbox{promptenv}[1]{
        boxrule = 1.5pt,
        fontupper = \small,
        fonttitle = \bf\color{black},
        arc = 5pt,
        rounded corners,
        colframe = black,
        colbacktitle = white!97!gray,
        colback = white!97!gray,
        title = #1,
}


\abstract{
Generalist foundation models (GFMs) are renowned for their exceptional capability and flexibility in effectively generalizing across diverse tasks and modalities.
In the field of medicine, while GFMs exhibit superior generalizability based on their extensive intrinsic knowledge as well as proficiency in instruction following and in-context learning, specialist models excel in precision due to their in-depth domain-specific knowledge.
In this work, for the first time, we explore the synergy between the GFM and specialist models, to enable precise medical image analysis on a broader scope.
Specifically, we propose a novel cooperative framework, \textbf{Generalist-Specialist Collaboration (GSCo)}, which consists of two stages, namely the construction of GFM and specialists, and collaborative inference on downstream tasks.
In the construction stage, we develop \textbf{\ours{}}, the largest open-source GFM tailored for medicine, showcasing exceptional instruction-following and in-context learning capabilities.
Meanwhile, a series of lightweight specialists are crafted for specific downstream tasks with low computational overhead.
In the collaborative inference stage, we introduce two cooperative mechanisms, Mixture-of-Expert Diagnosis (MoED) and Retrieval-Augmented Diagnosis (RAD), to harvest the generalist's in-context learning abilities alongside the specialists' domain expertise.
Concretely, MoED incorporates predictions from specialists as references, while RAD employs specialists to retrieve similar cases, collectively providing \ours{} with much more contextual guidance.
For a comprehensive evaluation, we curate a large-scale benchmark featuring 28 datasets and about 250,000 images across a wide range of medical modalities, including radiology, pathology, dermatology, ophthalmology, and gastroenterology.
Extensive experimental results demonstrate that \ours{} consistently outperforms state-of-the-art GFMs on downstream datasets.
Furthermore, GSCo exceeds both GFMs and specialists across all out-of-domain disease diagnosis datasets.
These findings indicate a significant paradigm shift in the clinical application of GFMs, transitioning from separate models for specific tasks to a collaborative approach between GFMs and specialists. 
This collaboration enables GFMs to perform precise medical image analysis even in out-of-domain scenarios, enhancing their scalability and sustainability, thereby advancing the frontiers of generalizable AI in medicine.
}


\maketitle
\section{Introduction}
Advanced by the rapid development of Large Language Models (LLMs)~\cite{achiam2023gpt4,touvron2023llama,anil2023palm} as well as vision-language pre-training~\cite{radford2021learning,tung2024exploring,xu2024multimodal}, large-scale vision-language models (LVLMs)~\cite{zhu2023minigpt,liu2024visual,chen2023internvl} have demonstrated remarkable performance in a wide range of tasks (e.g., visual question answering~\cite{antol2015vqa} and image captioning~\cite{lin2014microsoft}), thus establishing themselves as generalist foundation models (GFMs).
In the realm of medicine, GFMs~\cite{li2024llavamed, moor2023medflamingo, wu2023radfm, tu2024towards, moor2023foundation, zhou2024generalist, saab2024capabilities, yang2024advancing} also showcased impressive proficiency in generalizing across various tasks, such as visual question answering~\cite{he2020pathvqa, ImageCLEFVQA-Med2019} and radiology report generation tasks~\cite{johnson2019mimiccxr, demner2016iu, jin2024promptmrg, chen2024dia}.
The remarkable generalizability of GFMs can be attributed to two key aspects.
First, the extensive and diverse training corpus endows the models with comprehensive medical knowledge. 
Additionally, the powerful instruction following and in-context learning abilities of GFMs improve their versatility and flexibility, facilitating their applications across a multitude of tasks.
While GFMs exhibit superior generalizability, specialist models excel in precision. 
Tailored for specific downstream tasks, these specialist models possess profound domain-specific knowledge, enabling themselves to concentrate on a narrower scope and deliver more precise results.
For instance, in medical image diagnosis tasks~\cite{nguyen2021vindr, yang2023medmnist, wang2023real}, specialist models surpass the GFMs and demonstrate superior performance~\cite{tu2024towards, hu2024omnimedvqa, chen2024gmai}.
Therefore, exploring how to utilize the generalizability and flexibility of the GFMs, along with the expertise and precision of specialist models, represents a promising direction for generalizable AI in medicine.
\par
In this work, for the first time, we propose a novel cooperative framework, \textbf{Generalist-Specialist Collaboration (GSCo)}, to explore the synergy between the GFMs and specialist models.
\textbf{Fig.~\ref{fig:main} (a)} presents the overview of the GSCo framework, which consists of two stages: the construction of GFM and specialists, and collaborative inference on downstream tasks.
In the construction stage, a medical GFM, \textbf{\ours{}}, is developed based on a large-scale training corpus of medical image-text pairs across various modalities.
Meanwhile, a series of lightweight specialist models are tailored to specific downstream tasks with much lower computational consumption.
In the collaborative inference stage, two core mechanisms, namely \textbf{Mixture-of-Expert Diagnosis (MoED)} and \textbf{Retrieval-Augmented Diagnosis (RAD)}, are proposed to achieve cooperation between \ours{} and the specialist models.
\par
\begin{figure*}[!h]
\vspace{40pt}
\centering
\includegraphics[width=\linewidth, page=1]{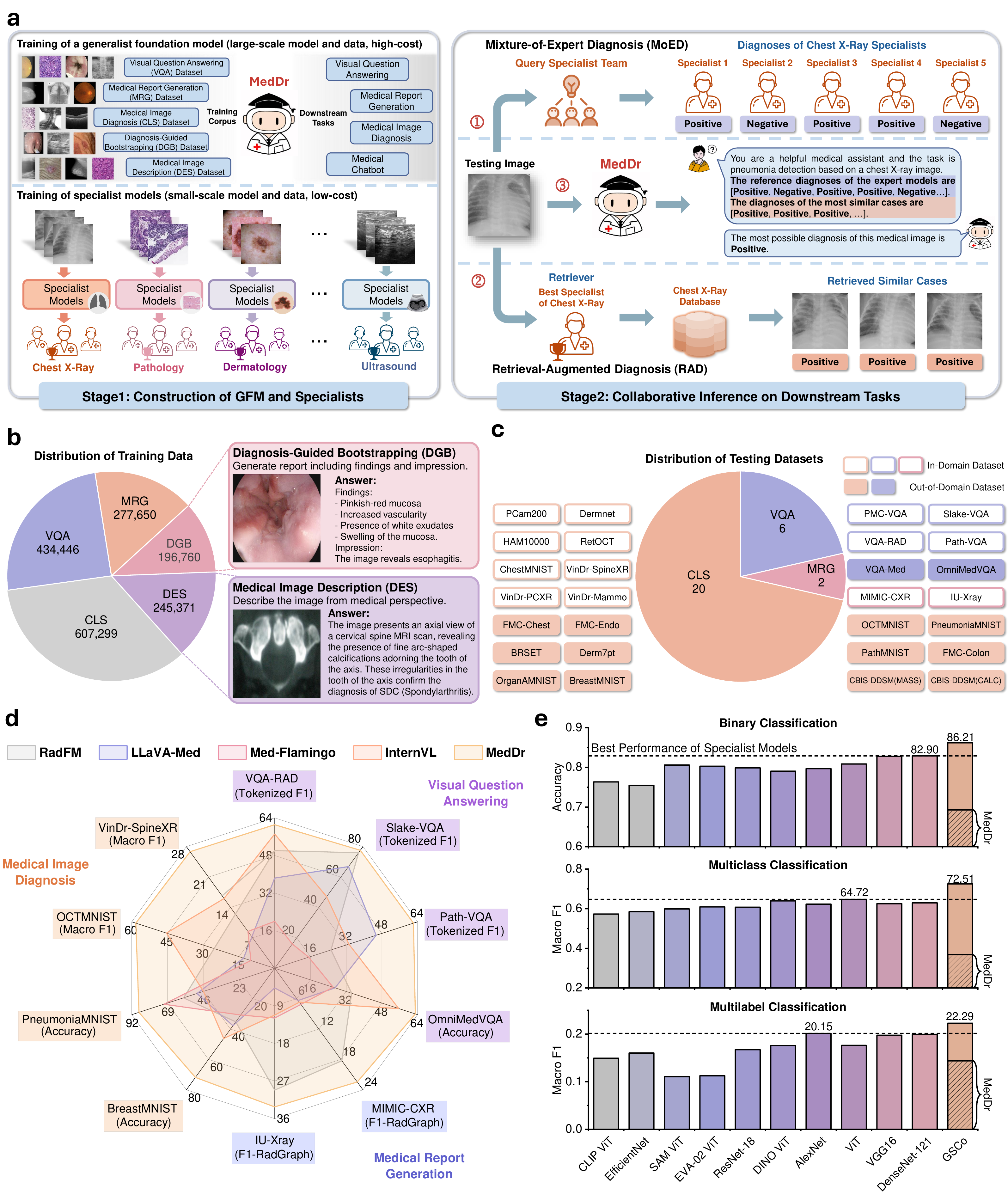}
\vspace{40pt}
\end{figure*}

\begin{figure*}[t]
\caption{
\textbf{Overview of the study.}
\textbf{(a)} Generalist-Specialist Collaboration (GSCo) Framework.
In the construction stage, based on a large-scale training corpus, \ours{} is developed, capable of performing diverse tasks and handling various medical modalities, demonstrating exceptional generalizability.  
Meanwhile, a series of specialist models are tailored for distinct tasks on specific datasets with low computational cost, excelling in expertise and precision.
In the collaborative inference stage, mixture-of-expert diagnosis (MoED) and retrieval-augmented diagnosis (RAD) are introduced to facilitate collaboration between \ours{} and specialist models.
MoED incorporates the diagnoses of specialist models as guidance, while RAD utilizes the specialist models to retrieve the most similar cases for reference.
The results of MoED and RAD are combined together as the context information fed into the \ours{}.
\textbf{(b)} Distribution of the training datasets. 
Two examples of the proposed Diagnosis-Guided Bootstrapping (DGB) and Medical Image Description (DES) data are presented.
\textbf{(c)} Distribution of the testing datasets.
The benchmark incorporates 14 in-domain datasets and 14 out-of-domain datasets.
\textbf{(d)} Performance of the GFMs.
\ours{} surpasses other GFMs, achieving state-of-the-art performance on diverse medical tasks across various medical modalities.
\textbf{(e)} Overall Performance on medical image diagnosis tasks.
GSCo excels through the synergy between \ours{} and specialist models.
}
\label{fig:main}
\end{figure*}
Specifically, in the construction stage, we focus on the development of an advanced medical GFM.
To curate a large-scale multi-modal training corpus, we introduce two novel datasets: the Diagnosis-Guided Bootstrapping (DGB) dataset and the Medical Image Description (DES) dataset.
The DGB dataset is constructed based on abundant medical image diagnosis data~\cite{pham2022vindr, tschandl2018ham10000, data8020029, wang2017chestx, pacheco2020pad, Smedsrud2021}, aiming to enhance the intrinsic disease diagnosis capabilities of the GFM.
In contrast to prior methods~\cite{li2024llavamed, wu2023radfm} that solely rely on textual information of image-text pairs to generate instruction-tuning data, the DGB dataset is generated by integrating both visual and textual information.
Concretely, we utilize the GFM~\cite{chen2023internvl} in the general domain to generate detailed medical reports, including findings and conclusions, based on diagnosis information (e.g., classification labels).
Guided by these human-verified annotations, the generated data not only demonstrates increased reliability but also significantly enriches the depth and diversity of visual information conveyed in the text.
Additionally, the DES dataset is presented to broaden the scope of the training data, incorporating image-based case studies corresponding to diverse medical conditions from OpenI~\cite{demner2012design}.
We also employ a GFM~\cite{chen2023internvl} to rewrite the text of the case study and remove the information that cannot be derived from the corresponding image.
In addition to DGB and DES datasets, we incorporate existing Medical Image Diagnosis (CLS)~\cite{tschandl2018ham10000}, Medical Report Generation (MRG)~\cite{johnson2019mimiccxr}, and Visual Question Answering (VQA)~\cite{lau2018visual} datasets as well.
Overall, the training corpus consists of more than 2 million samples across five distinct types of training datasets, covering a wide range of medical modalities. 
\textbf{Fig.~\ref{fig:main} (b)} presents the distribution of the training corpus and two samples from the DGB and DES datasets respectively.
\par
Building upon the training corpus, we develop \textbf{\ours{}}, the largest open-source generalist foundation model for medicine consisting of 40B parameters. 
Compared to other generalist foundation models for medicine~\cite{li2024llavamed, moor2023medflamingo, wu2023radfm}, \ours{} exhibits superior capabilities in medical image analysis across more diverse medical modalities, including radiology, pathology, dermatology, ophthalmology, and gastroenterology.
Moreover, \ours{} demonstrates advanced in-context learning and instruction-following capability, fostering the collaboration between the generalist foundation model and the specialist models.
\par
In the collaborative inference stage of the GSCo framework, we delve into the synergistic relationship between \ours{} and specialist models.
To exploit the generalist’s in-context learning abilities alongside the specialists’ domain expertise, mixture-of-expert diagnosis (MoED), and retrieval-augmented diagnosis (RAD) are proposed as the core mechanisms of the collaboration between GFMs and specialists.
Mixture-of-expert diagnosis aims to boost the generalist foundation model with the prediction results of specialist models (\textbf{Fig.~\ref{fig:main} (a)} and \textbf{Fig.~\ref{fig:moed_method}}). 
The paradigm of mixture-of-expert~\cite{jacobs1991adaptive}, which ensembles the insights from multiple experts, has been explored to enhance the robustness of the predictions~\cite{fedus2022switch, luo2024towards, xiong2024mome}.
In MoED, the outputs of the specialist models act as the reference context and are fed into \ours{} with the testing image together.
Then, \ours{} is required to provide the diagnosis considering both the content of the testing image and the results provided by the specialist models.
Different from MoED, which exploits the inherent expert knowledge of the specialist model, retrieval-augmented diagnosis is proposed to fully leverage the broad medical knowledge embedded within the existing data (\textbf{Fig.~\ref{fig:main} (a)} and \textbf{Fig.~\ref{fig:rad_method}}).
In contrast to previous medical generalist foundation models~\cite{li2024llavamed, moor2023medflamingo, wu2023radfm, tu2024towards} that relied solely on the internal knowledge of the model to make diagnoses, we incorporate Retrieval-Augmented Generation (RAG)~\cite{lewis2020retrieval} to leverage external knowledge, thereby enhancing model accuracy and reliability.
Concretely, in the proposed RAD mechanism, each specialist model serves as a retriever, using the visual embedding of the testing image as the query to retrieve the most similar samples in the database.
The information from the retrieved samples is then provided to the generalist foundation model as a contextual reference to assist medical image analysis.
MoED and RAD collectively provide helpful guidance to \ours{} based on the expertise of the specialists.
Meanwhile, as a decision-maker, \ours{} integrates its intrinsic knowledge with external knowledge to render the final diagnosis.
\par
To comprehensively evaluate \ours{} and the proposed GSCo framework, we curate a large-scale benchmark.
Compared with previous works~\cite{li2024llavamed, moor2023medflamingo, wu2023radfm, tu2024towards, zhang2024generalist}, this benchmark excels in both diversity and magnitude.
As shown in \textbf{Fig.~\ref{fig:main} (c)}, it encompasses diverse medical datasets such as medical image diagnosis, visual question answering, medical report generation tasks, etc.
Specifically, there are 14 in-domain datasets and 14 out-of-domain datasets, resulting in a total of 250,000 samples. 
Notably, to assess the improvements brought by GSCo to GFM on medical image diagnosis datasets, where specialists often yield superior results due to their domain-specific knowledge, we integrate 20 distinct medical image diagnosis datasets, including 11 medical modalities. 
\par
We first conduct experiments on GFMs to demonstrate the superiority of \ours{}.
\textbf{Fig.~\ref{fig:main} (d)} presents the comparison among GFMs on 10 benchmark datasets that span different medical tasks and modalities.
\ours{} consistently surpasses other GFMs from both medical and general domains by a large margin.
We then perform experiments to validate the effectiveness of GSCo.
\textbf{Fig.~\ref{fig:main} (e)} illustrates performance on medical image diagnosis tasks.
Despite the competitive performance of specialist models due to their domain-specific knowledge, the proposed GSCo framework further improves the performance of \ours{} and surpasses all specialist models.
These experimental results highlight the significance of the proposed GSCo framework, representing a paradigm shift in the clinical application of GFMs.  
This transition moves from utilizing separate models for medical tasks independently to fostering collaboration between GFMs and specialist models.
The advantages of the GSCo framework are twofold:
Firstly, GSCo is effective.
Compared with the independent use of either GFMs or specialist models, GSCo demonstrates superior performance, particularly on out-of-domain datasets, showcasing its advanced generalizability.
Secondly, GSCo is efficient. 
When confronted with out-of-domain tasks or data, rather than investing substantial resources to fine-tune the GFM, it can efficiently adapt lightweight specialist models with minimal consumption, indicating its scalability and sustainability.
\par
To summarize, our contributions are as follows:
\begin{itemize}
\item We introduce Generalist-Specialist Collaboration (GSCo), the first collaborative framework to explore the synergy between the GFM and specialist models. 
GSCo harvests the generalist’s in-context learning ability and the specialists’ domain expertise, to enable precise medical image analysis on diverse medical tasks.
This synergistic paradigm not only broadens the functionalities of the GFM with efficient resource utilization but also ensures scalability and sustainability, thereby catalyzing the advancement of generalizable AI in the medical field.

\item We present \ours{}, the largest open-source generalist foundation model tailored for medicine. 
Concretely, in the development of \ours{}, diagnosis-guided bootstrapping (DGB) and medical image description (DES) are introduced to enhance the diversity of the training corpus.
As a result, \ours{} can handle various medical modalities and tasks, achieving state-of-the-art performance in downstream tasks and outperforming other GFMs.
Additionally, \ours{} excels in instruction-following and in-context learning, providing a better foundation for collaboration with specialist models.

\item We propose two cooperative mechanisms, mixture-of-expert diagnosis (MoED) and retrieval-augmented diagnosis (RAD), to facilitate collaboration between \ours{} and specialist models.
MoED incorporates the diagnoses of specialists as guidance, while RAD utilizes the specialists to retrieve the most similar cases for reference.
The results of MoED and RAD are combined together as the context information fed into the \ours{} as guidance.

\item 
We establish a large-scale benchmark, which comprises 28 datasets with about 250,000 test images, covering more than ten medical modalities across various medical tasks.
Extensive experiments are conducted on the benchmark, demonstrating the superior capabilities of \ours{} and validating the efficacy of the proposed GSCo framework.
\end{itemize}

\section{Results}
\subsection*{Comprehensive evaluation benchmark in medicine}
In this study, to thoroughly assess the model's performance of the medical tasks, we curated a large-scale benchmark, encompassing 28 public datasets comprising about 250,000 images.
Compared with previous work~\cite{li2024llavamed, moor2023medflamingo, wu2023radfm, tu2024towards, zhang2024generalist}, this benchmark excels in both diversity and magnitude.
The datasets within our benchmark are carefully selected to include clinically pertinent tasks such as medical image diagnosis, visual question answering, and medical report generation. 
Additionally, the benchmark spans a diverse range of medical image modalities, including radiology, pathology, dermatology, ophthalmology, gastroenterology, etc., ensuring a comprehensive evaluation of the model's capabilities across various medical conditions.
\paragraph{\textup{Medical Image Diagnosis}}
Medical Image Diagnosis is one of the most fundamental tasks in medicine, which requires the model to diagnose the queried images within a predefined label set.
In our benchmark, we integrate 20 distinct medical image diagnosis datasets, encompassing approximately 100,000 testing samples and 11 medical modalities.
This benchmark has two prominent features.
Firstly, the benchmark covers diverse medical conditions. 
For instance, VinDr-SpineXR~\cite{nguyen2021vindr} and FMC-Chest~\cite{wang2023real} are challenging radiology datasets that focus on spine and chest X-ray images, respectively.
HAM10000~\cite{tschandl2018ham10000} and DermNet~\cite{kaggleDermnet} are both datasets related to skin diseases, with the former focusing more on dermatoscopic images and the latter more on clinical images.
RetOCT~\cite{9740985} and BRSET~\cite{nakayama2023brazilian} focus on ocular diseases despite with different medical modalities.
Secondly, the benchmark incorporates different classification types.
For example, PneumoniaMNIST~\cite{yang2023medmnist} and BreastMNIST~\cite{yang2023medmnist} are binary classification datasets. 
While providing an available answer (i.e., Positive or Negative) for a binary classification task is straightforward for a GFM, achieving accurate predictions is indeed challenging.
FMC-Endo~\cite{wang2023real} and OCTMNIST~\cite{yang2023medmnist} are multi-class classification datasets with label sets consisting of 5 and 4 labels, respectively.
For each given image, the model is required to predict only one label among the predefined label sets.
The most challenging task is multi-label classification, where each test image may possess multiple, or even no, labels.
For example, ChestMNIST~\cite{wang2017chestx} is a multi-label classification dataset consisting of 15 categories, with each sample potentially having one or more positive labels.
Such tasks impose greater demands on the model's disease diagnosis and task understanding capabilities than binary and multi-class classification tasks.
To facilitate comparative analysis, we categorize these medical image diagnosis datasets into two groups: in-domain datasets and out-of-domain datasets, based on whether the training split of the dataset is included in the training corpus of \ours{}.
Extended Data Table~\ref{tab:indomain_data_stat} and Extended Data Table~\ref{tab:outdomain_data_stat} illustrate more detailed information about these datasets.
For evaluation metrics, we employ accuracy and macro-F1 for these medical image diagnosis tasks.
For binary classification datasets, the results are presented in terms of accuracy.
For multi-class and multi-label classification datasets, due to the unbalanced distribution of samples across different classes, the results are reported in terms of the Macro-F1 score.
Further details regarding the metric employed can be found in Section~\ref{sec:metric}.
\paragraph{\textup{Visual Question Answering}}
Visual Question Answering (VQA) task requires the model to answer the question based on the given image, requiring a thorough comprehension of both the visual and textual information.
We conduct experiments on 6 distinct VQA datasets, comprising 4 in-domain datasets: VQA-RAD~\cite{lau2018visual}, Slake-VQA~\cite{liu2021slake}, Path-VQA~\cite{he2020pathvqa}, and PMC-VQA~\cite{zhang2023pmcvqa}, as well as 2 out-of-domain datasets: VQA-Med~\cite{ImageCLEFVQA-Med2019} and OmniMedVQA~\cite{hu2024omnimedvqa}. 
The diversity inherent in these benchmark datasets allows us to extract valuable insights across various domains and facilitates a thorough evaluation of model performance.
For example, VQA-RAD, Slake-VQA, and VQA-Med datasets primarily focus on radiology data, such as CT, MRI, and X-ray images, while Path-VQA is mainly about pathology data. 
In contrast, the PMC-VQA and OmniMedVQA datasets are larger in scale and cover a broader spectrum of medical modalities.
PMC-VQA is built on PubMed while OmniMedVQA is derived from a wide range of medical datasets.
Notably, aside from OmniMedVQA, which consists of multiple-choice questions, the other datasets contain free-form questions.
For the evaluation of OmniMedVQA, we employ accuracy as our metric.
For other datasets, following MultiMedEval~\cite{royer2024multimedeval}, we evaluate the results using both natural language generation (NLG) and classification metrics.
\paragraph{\textup{Medical Report Generation}}
Medical Report Generation (MRG) involves the model's ability to enumerate all observations and deliver a diagnosis based on the analysis of the medical image. 
This task presents a significant challenge, as it requires the model to accurately capture the complexities inherent in the medical images being evaluated.
To assess the performance of MRG, we conduct experiments on two benchmark datasets, i.e., MIMIC-CXR~\cite{johnson2019mimiccxr} and IU-Xray~\cite{demner2016iu}. 
Both datasets focus on chest X-ray images and offer detailed medical reports that summarize patient conditions.
For the evaluation of the models, we utilize both NLG metrics and model-based metrics in MultiMedEval~\cite{royer2024multimedeval}.
Further details regarding the metric employed in VQA and MRG tasks can be found in Section~\ref{sec:metric}.
\subsection*{\ours{} demonstrates superior performance in medical image diagnosis}
\begin{figure*}[!h]
\centering
\vspace{5pt}
\includegraphics[width=\linewidth, page=1]{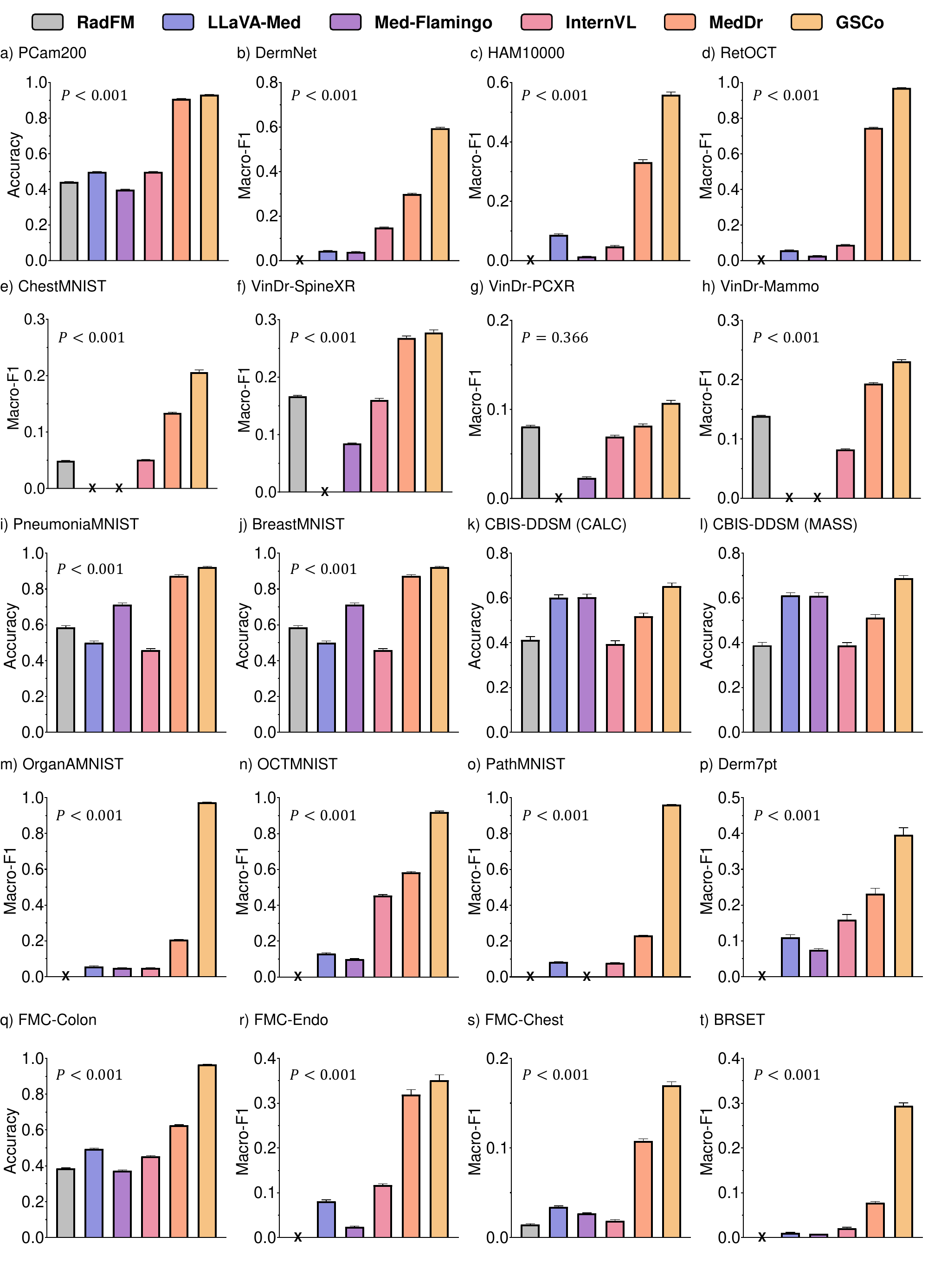}
\end{figure*}

\begin{figure*}[t]
\caption{
\textbf{Experimental results of different GFMs and the GSCo framework on medical image diagnosis datasets.}
(a)-(h): In-domain datasets. 
(i)-(t): Out-of-domain datasets.
For binary classification datasets (e.g., PCam200), the results are reported in terms of Accuracy. For multi-class (e.g., DermNet) and multi-label datasets (e.g., ChestMNIST), the results are presented in terms of the Macro-F1 score.
If the performance of \ours{} is the best one compared to other GFMs, P-value would be presented.
The model marked with ``\ding{53}'' cannot generate an appropriate response to evaluate its performance.
Detailed results are presented in \textbf{Extended Data Tables~\ref{tab:cls_binary}}, \textbf{\ref{tab:cls_multiclass}} and \textbf{\ref{tab:cls_multilabel}}.
\label{fig:cls_res}
}
\end{figure*}
We first explore the performance of the GFMs on medical image diagnosis tasks.
It should be noted that some GFMs are unable to produce appropriate responses used for evaluation on specific datasets, so we do not include them in the comparison.
If the performance of \ours{} is the best one compared to other GFMs, the P-value would be presented.
The experimental results are illustrated in \textbf{Fig.\ref{fig:cls_res}}. Specifically, \textbf{Figs.\ref{fig:cls_res}(a)-(h)} represent the in-domain results, while \textbf{Figs.\ref{fig:cls_res}(i)-(t)} depict the out-of-domain results.
For binary classification datasets (e.g., PCam200), the results are reported in terms of accuracy. For multi-class (e.g., DermNet) and multi-label datasets (e.g., ChestMNIST), the results are presented in terms of macro-F1 score.
\par
Overall, \ours{} outperforms other GFMs significantly and demonstrates its superiority in three aspects.
Firstly, \ours{} is proficiency in instruction following.
We note that RadFM, LLaVA-Med, and Med-Flamingo occasionally struggle to generate appropriate outputs for tasks that involve multiple labels. 
For instance, in multi-class classification tasks where only one label exists, they may either output the entire label set or merely rephrase the instructions, suggesting their inferior instruction following capabilities.
In contrast, \ours{} effectively adheres to instructions across a diverse range of tasks, consistently generating coherent and contextually appropriate outputs.
Secondly, \ours{} can handle a broader scope of medical modalities.
RadFM~\cite{wu2023radfm} is a GFM focusing on the radiology data so that it achieves commendable performance on radiology datasets such as VinDr-SpineXR~\cite{nguyen2021vindr} and VinDr-PCXR~\cite{pham2022vindr}.
However, its performance diminishes when applied to other medical modalities, indicating its limited generalization capabilities.
In comparison, \ours{} is capable of processing diverse medical modalities, including radiology, pathology, dermatology, ophthalmology, gastroenterology, etc.
Thirdly, \ours{} excels in medical image diagnosis.
LLaVA-Med~\cite{li2024llavamed} and Med-Flamingo~\cite{moor2023medflamingo} are mainly trained on visual question answering datasets. 
Their overall performance in medical image diagnosis tasks is still far from satisfactory.
Meanwhile, InternVL~\cite{chen2023internvl} can follow the instructions well, but its performance still lags \ours{} by a large margin due to its limited inherent medical knowledge.
For instance, on the FMC-Endo~\cite{wang2023real} dataset, \ours{} obtains a 32.0\% Macro-F1 score and significantly outperforms InternVL (11.7\% Macro-F1 score, $P<0.001$, \textbf{Fig.~\ref{fig:cls_res}(r)}).
Notably, in some binary classification datasets, such as CBIS-DDSM (CALC)~\cite{sawyer2016curated} and CBIS-DDSM (MASS)~\cite{sawyer2016curated}, LLaVA-Med, Med-Flamingo and InternVL struggle to distinguish between the two classes, consistently assigning the same label to all samples.
Although these models may achieve higher accuracy, their F1 score is 0.
This lack of differentiation indicates that these models may fail to capture the pertinent features necessary for effective classification, thereby undermining the reliability of their results.
In contrast, \ours{} achieves outstanding performance across various medical modalities and classification tasks, thereby highlighting its advanced capabilities in medical image analysis.

\subsection*{\ours{} excels in visual question answering and medical report generation}
\begin{figure*}[!h]
\centering
\includegraphics[width=0.99\linewidth, page=1]{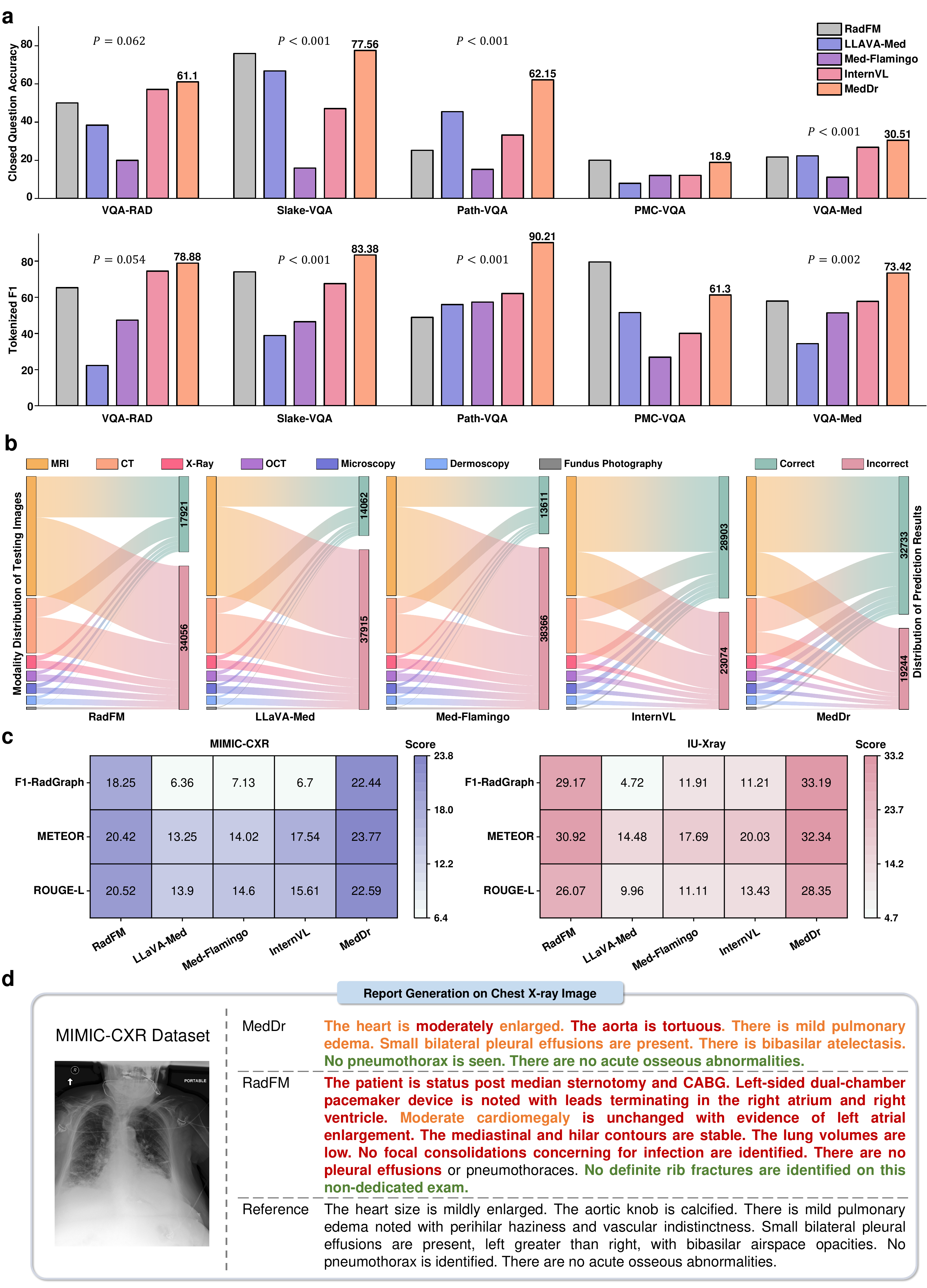}
\vspace{10pt}
\end{figure*}

\begin{figure*}[!h]
\caption{
\textbf{Experimental results of GFMs on medical visual question answering and medical report generation tasks.}
\textbf{(a)} Results in terms of Closed Question Accuracy and Tokenized F1-Score on VQA-RAD, Slake-VQA, Path-VQA, PMC-VQA, and VQA-Med datasets.
If the performance of \ours{} is the best one compared to other GFMs, P-value would be presented.
\textbf{(b)} Medical modality and result distribution of the generalist foundation models on the OmnimedVQA dataset. The number refers to the quantity of multiple-choice questions.
\textbf{(c)} Results in terms of ROUGE-L, METEOR, F1-RadGraph on medical report generation datasets MIMIC-CXR and IU-Xray. 
\textbf{(d)} An example of medical report generation on chest X-ray images.
The text in \textbf{\textcolor{green1}{Green}} indicates the correct normal contents. 
The text in \textbf{\textcolor{orange1}{Orange}} indicates the correct abnormal contents.
The text in \textbf{\textcolor{red1}{Red}} indicates the incorrect contents.
Reference denotes the ground truth of the image from the corresponding dataset.
Results are accessed by a qualified radiologist.
Detailed results are presented in \textbf{Extended Data Tables~\ref{tab:indomainvqa},\ref{tab:outofdomainvqa},\ref{tab:omnimed_vqa}, and~\ref{tab:mrg}}.
}
\label{fig:vqa_mrg}
\end{figure*}
Here, we delve into the evaluation of GFMs in visual question answering and medical report generation tasks.
The results of VQA tasks are presented in \textbf{Fig.~\ref{fig:vqa_mrg}(a)-(b)}, \textbf{Extended Data Table~\ref{tab:indomainvqa}} and \textbf{Extended Data Table~\ref{tab:outofdomainvqa}}.
\par
Overall, \ours{} consistently outperforms other GFMs across all datasets.
Concretely, on the VQA-RAD dataset, \ours{} achieves a 59.62\% BLEU-1 score ($P=0.062$) and a 61.10\% F1 score ($P=0.054$), much better than the fine-tuned LLaVA-Med.
On the Slake-VQA dataset, \ours{} obtains an 83.38\% accuracy on close-ended questions and a 77.26\% recall, surpassing RadFM by a large margin ($P<0.001$, \textbf{Fig.~\ref{fig:vqa_mrg}(a)}). 
For the most challenging dataset PMC-VQA, \ours{} achieves a 27.30\% recall and a 14.94\% accuracy on open questions, outperforming other models significantly ($P<0.001$, \textbf{Extended Data Table~\ref{tab:indomainvqa}}).
On the OmniMedVQA datasets, as illustrated in \textbf{Fig.~\ref{fig:vqa_mrg}(b)} and \textbf{Extended Data Table~\ref{tab:omnimed_vqa}}, \ours{} also establishes superior performance across all medical modalities, resulting in 63.0\% overall accuracy.
The overall performance of \ours{} in the visual question answering task demonstrates that the model not only excels in comprehending both visual and textual information but also handles a greater diversity of medical modalities.
\par
The results of MRG tasks are presented in \textbf{Fig.~\ref{fig:vqa_mrg}(c)} and \textbf{Extended Data Table~\ref{tab:mrg}}.
It should be noted that aside from RadFM~\cite{wu2023radfm} and \ours{}, other models such as LLaVA-Med~\cite{li2024llavamed}, Med-Flamingo~\cite{moor2023medflamingo}, and InternVL~\cite{chen2023internvl} have not been trained on datasets of MRG tasks. 
\ours{} outperforms RadFM~\cite{wu2023radfm} (overall, $P<0.001$), which mainly focuses on radiology tasks, across nearly all evaluated metrics on both benchmark datasets.
For example, \ours{} achieves a ROUGE-L score of 22.59 on the MIMIC-CXR dataset and 28.35 on the IU-Xray dataset, highlighting its proficiency in comprehensively interpreting medical images.
\par
\textbf{Fig.~\ref{fig:vqa_mrg}(d)} presents a challenging case of medical report generation on a chest X-ray image, characterized by the presence of multiple abnormalities in the patient.
We compare \ours{} with RadFM~\cite{wu2023radfm}, which is a GFM specialized in radiology.
The outputs from both \ours{} and RadFM are evaluated by a qualified radiologist.
Notably, \ours{} demonstrates superior performance by accurately identifying most abnormalities, including ``mild pulmonary edema'', ``small bilateral pleural effusions'', and ``bibasilar atelectasis'', thereby underscoring its advanced capabilities in the domain of medical image analysis.
\subsection*{GSCo enables accurate disease diagnosis and generalizable AI for medicine}
In this section, we conduct experiments on medical image diagnosis datasets to demonstrate the effectiveness of the proposed Generalist-Specialist Collaboration mechanism.
Following previous method~\cite{doerrich2024rethinking}, we select ten representative models in computer vision (e.g., ResNet~\cite{he2016deep} and ViT~\cite{dosovitskiy2020image}) as the foundation models, which possess significantly fewer parameters compared to the GFM.
Details regarding the selected models can be found in Section~\ref{sec:gen_spe_model}.
We fine-tune these foundation models on each of the 20 medical image diagnosis datasets, resulting in a total of 200 specialist models.
Moreover, for a fair comparison, we introduce a baseline method, ``Voting'', where results are aggregated and voted from the predictions of the specialists, effectively functioning as a naive collaborative method to exploit the results of the specialist models.
\paragraph{\textup{Quantitative Analysis}}
\par
\begin{figure}[!ht]
\centering
\includegraphics[width=\linewidth, page=1]{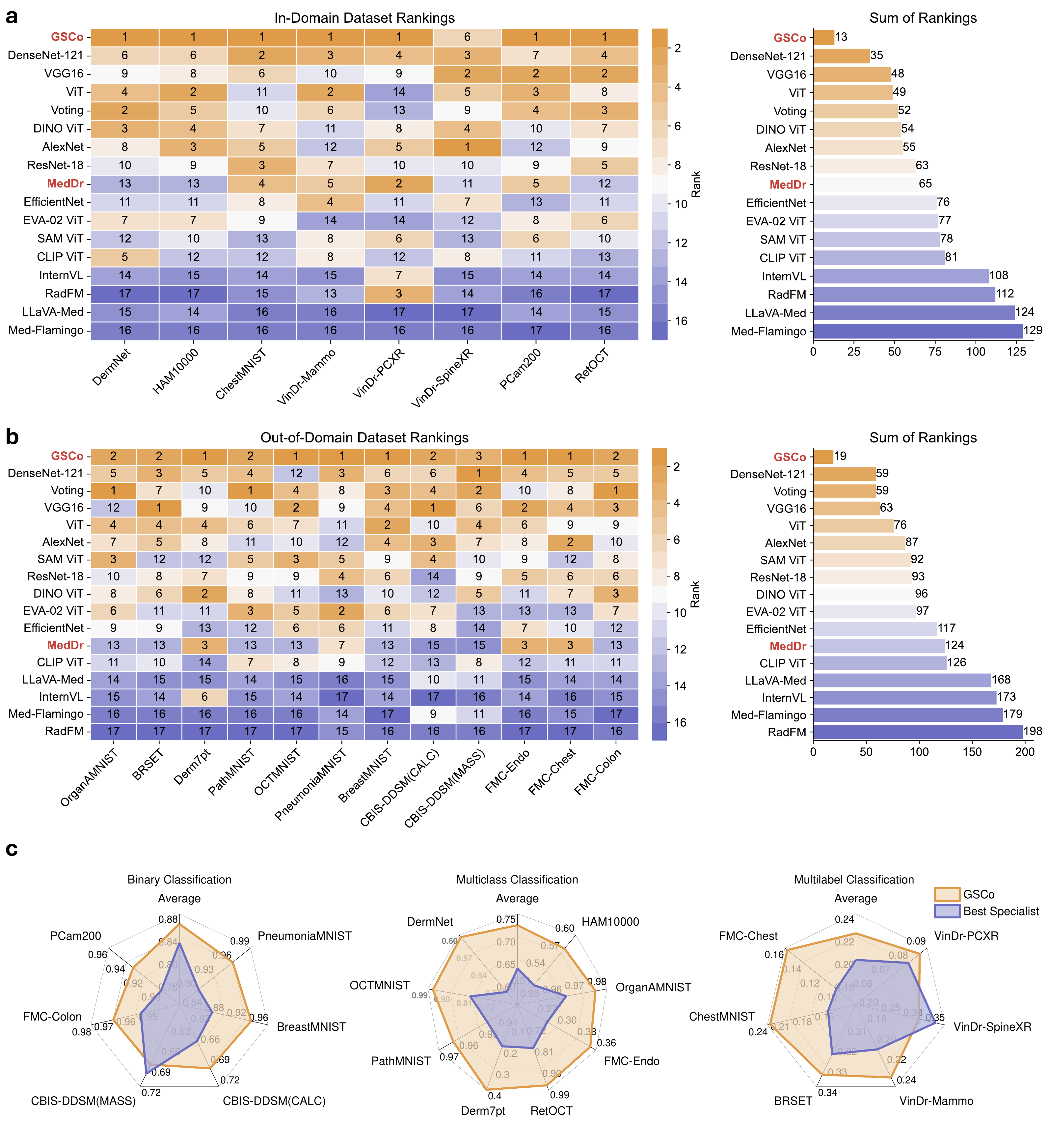}
\caption{
\textbf{Overall experimental results on medical image diagnosis datasets.}
\textbf{(a)} Overall ranking of different models on 8 in-domain datasets and the summation of the rankings. 
The model with better performance is assigned a lower ranking number.
\textbf{(b)} Overall ranking of different models on 12 out-of-domain datasets and the summation of the rankings. 
\textbf{(c)} Comparison between the best specialist model DenseNet-121 and the proposed framework GSCo.
The datasets are divided into three groups according to the classification types. 
For binary classification, accuracy is reported. 
Otherwise, the Macro-F1 score is reported.
The detailed experimental results are presented in \textbf{Extended Data Tables~\ref{tab:cls_binary}, \ref{tab:cls_multiclass} and \ref{tab:cls_multilabel}}.
}
\label{fig:gsc_res}
\vspace{10pt}
\end{figure}
\textbf{Fig.~\ref{fig:gsc_res}(a)} and \textbf{Fig.~\ref{fig:gsc_res}(b)} illustrate the ranking order of various methods on the in-domain datasets and out-of-domain datasets, respectively.
The summations of their rankings across different tasks are also presented.
Firstly, we observe that \ours{} outperforms other GFMs and even surpasses some specialist models on in-domain datasets.
Compared with specialist models, the overall performance of GFMs is lower, with most GFMs ranking relatively poorly.
This observation indicates that while GFMs demonstrate superior generalizability through performing diverse tasks with a unified model, specialist models excel in precision on specific datasets due to their domain-specific fine-tuning.
Notably, on in-domain datasets, \ours{} showcases superior performance compared to several specialist models.
For instance, on the ChestMNIST and PCam200 datasets, \ours{} surpasses the majority of specialized models, achieving 4th and 5th place, respectively, which highlights its exceptional intrinsic diagnostic capabilities.
Secondly, we find that GSCo achieves the highest overall performance, significantly surpassing other methods.
As a straightforward collaborative method, ``Voting'' shows improvements on most datasets when compared with specialist models, highlighting its effectiveness.
However, ``Voting'' obtains superior performance on binary classification and multi-class classification datasets while acquiring inferior results on multi-label datasets.
This discrepancy suggests that it may struggle to effectively leverage the prediction results of specialist models on multi-label classification tasks.
This is because, as a naive method, ``Voting'' relies solely on the outputs of the specialist models and treats their suggestions equally.
Consequently, when confronted with highly challenging multi-label classification tasks, if the majority of specialized models provide incorrect predictions, ``Voting'' may yield erroneous diagnoses.
In contrast, the proposed GSCo framework not only considers the predictions from the specialist models but also leverages the inherent knowledge of \ours{}, leading to superior performance and robustness.
\textbf{Fig.~\ref{fig:gsc_res}(c)} presents the comparison between the specialist model with the best overall performance, DenseNet-121, and the proposed framework GSCo.
For clarity, the results are categorized into three groups based on their classification task types.
Compared with the best specialist model, GSCo exhibits a notable performance advantage, even on out-of-domain datasets, underscoring its superiority and generalizability.
\par
In conclusion, the Generalist-Specialist Collaboration framework exemplifies a synergistic relationship between GFM and specialist models.
On the one hand, specialist models, with their domain-specific knowledge, offer guidance to GFM, thereby significantly enhancing its performance, especially on out-of-domain datasets.
Additionally, fine-tuning the specialist model on specific downstream datasets is computationally efficient and can be implemented with minimal additional resources.
On the other hand, the GFM acts as decision-makers with extensive intrinsic medical knowledge.
Different from the ``Voting'' method, which relies solely on the outputs of specialist models, \ours{} retains its diagnostic capabilities while integrating the guidance from these specialists. 
This collaborative strategy effectively bolsters performance across a wide range of medical tasks.
\begin{figure}[!t]
\centering
\vspace{30pt}
\includegraphics[width=\linewidth, page=1]{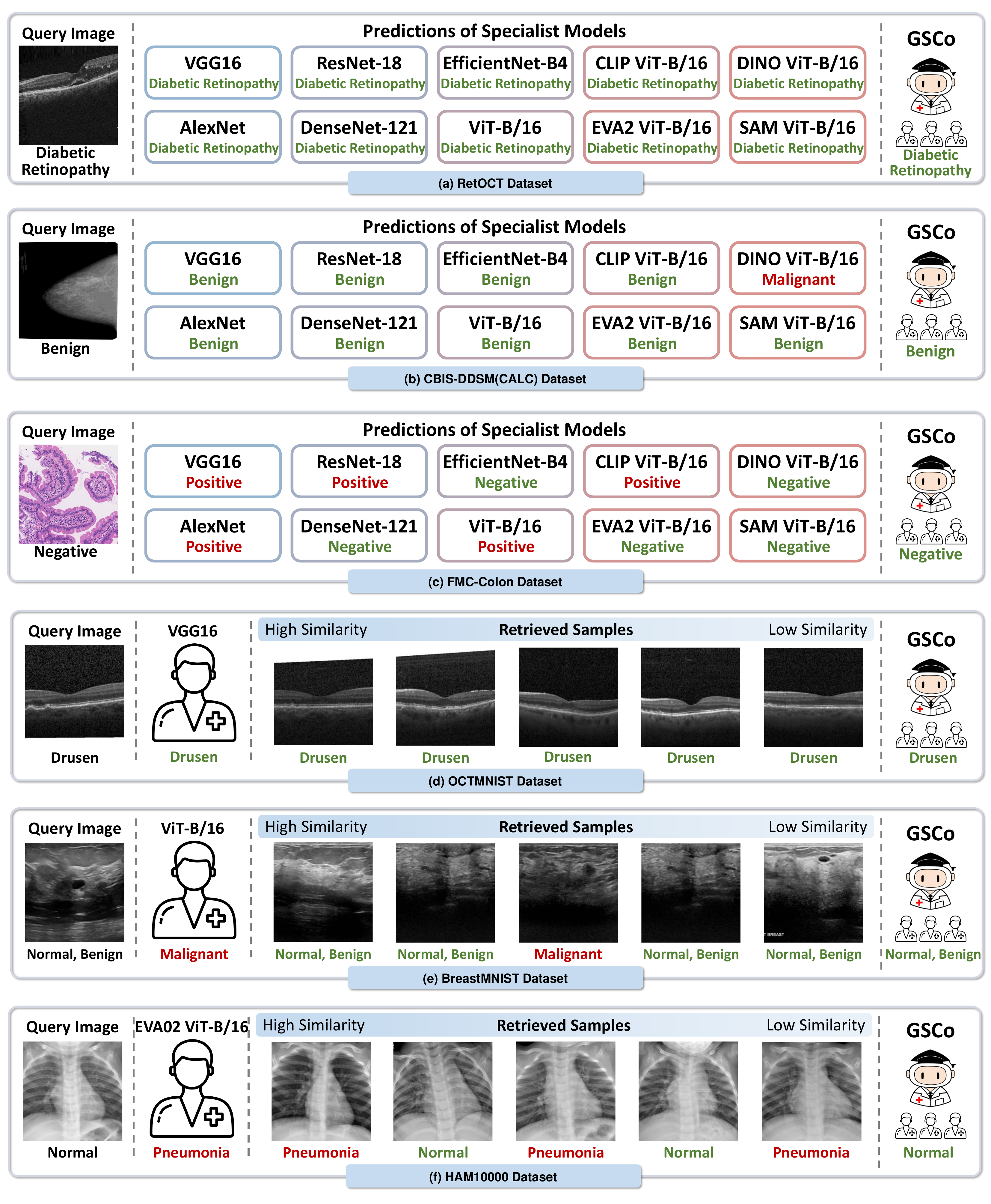}
\vspace{30pt}
\end{figure}

\begin{figure}[!t]
\caption{
Examples of mixture-of-expert diagnosis and retrieval-augmented diagnosis on downstream medical image diagnosis datasets.
(a)-(c): Examples of mixture-of-expert diagnosis. Query image and its label, predictions of specialist models, and prediction of GSCo are listed.
(d)-(e): Examples of retrieval-augmented diagnosis. Query image and its label, best specialist model and its prediction, top-$5$ retrieved similar images and their labels, and prediction of GSCo are listed.
\textbf{\textcolor{green1}{Green}} indicates the correct results while \textbf{\textcolor{red1}{Red}} indicates the erroneous results.
More examples are presented in \textbf{Extended Data Figures~\ref{fig:moed_res_more} and~\ref{fig:rad_res_more}}.
}
\label{fig:moed_rad_res}
\end{figure}
\paragraph{\textup{Qualitative Analysis}}
We visualize the experimental results of mixture-of-expert diagnosis (MoED) and retrieval-augmented diagnosis (RAD).
\textbf{Fig.~\ref{fig:moed_rad_res}(a)-(c)} and \textbf{Extended Data Fig.~\ref{fig:moed_res_more}} showcase examples of the MoED on downstream datasets across various medical modalities.
Firstly, we find that the aggregation of predictions from multiple specialist models yields accurate and robust guidance.
In medical image diagnosis, the key to accurately identifying a disease often lies in the subtle nuances present within the image.
Due to this inherent difficulty, among all specialist models, only EfficientNet-B4~\cite{tan2019efficientnet} consistently produces correct diagnostic results across all example cases.
Meanwhile, the aggregation of predictions from multiple specialist models, such as ``Voting'', can derive more accurate results.
For instance, on the RetOCT dataset (\textbf{Fig.~\ref{fig:moed_rad_res} (a)}) and CBIS-DDSM(CALC) dataset (\textbf{Fig.~\ref{fig:moed_rad_res} (b)}), the majority of specialist models make correct predictions, thereby providing \ours{} with effective guidance to derive accurate diagnoses.
Secondly, we note that even if the aggregating results fail to offer correct reference, \ours{} can arrive at the correct diagnoses as well.
For example, on the FMC-Colon dataset (\textbf{Fig.~\ref{fig:moed_rad_res} (c)}), half of the specialist models predict ``Positive'', while the other half provide opposite results.
While the naive ``Voting'' strategy fails in such a dilemma, \ours{} generates the correct result ``Negative''.
These observations validate the superiority of \ours{} and the efficacy of the proposed MoED.
\ours{} not only leverages the reference diagnoses provided by the specialist models but also utilizes its inherent knowledge to make the final decision.
MoED represents an effective collaboration that integrates the strengths of both the Generalist and specialist models, thereby enhancing diagnostic accuracy.
\par
\textbf{Fig.~\ref{fig:moed_rad_res}(d)-(f)} and \textbf{Extended Data Fig.~\ref{fig:rad_res_more}} present examples of the RAD on downstream datasets covering a broad range of modalities.
Firstly, we delve into the effectiveness of the retrieval strategy.
As illustrated in \textbf{Fig.~\ref{fig:moed_rad_res} (d)} and \textbf{Extended Data Fig.~\ref{fig:rad_res_more} (e)}, the retrieved images share the same label as the query image, thereby serving as reliable references.
Notably, although the specialist may render an incorrect diagnosis as a predictor, the majority of images retrieved still provide accurate diagnostic information, demonstrating the specialist's robust capability as a retriever.
For instance, in \textbf{Fig.~\ref{fig:moed_rad_res} (e)}, the specialist model predicts the ``Malignant'' while most of the retrieved samples are ``Normal, Benign''.
These observations suggest that, in most cases, the retrieved samples can offer accurate guidance for \ours{}, thereby validating the precision and robustness of the retrieval strategy.
Secondly, we explore the efficacy of RAD.
In most cases of \textbf{Fig.~\ref{fig:moed_rad_res}(d)-(f)} and \textbf{Extended Data Fig.~\ref{fig:rad_res_more}}, the retrieved samples provide correct guidance.
Meanwhile, we observe that even if the predictions from the specialist model and the retrieved items contain distracting information, \ours{} can still make an accurate diagnosis based on its inherent disease diagnosis capability.
For example, as shown in \textbf{Fig.~\ref{fig:moed_rad_res} (f)}, while most retrieved images are "Pneumonia" and therefore provide erroneous guidance, \ours{} successfully derive the ``Normal'' diagnosis.
These findings underscore that \ours{} adeptly harnesses not only the external knowledge provided by the retrieved samples but also its inherent diagnostic capabilities, thereby demonstrating the effectiveness of the proposed RAD approach.
\paragraph{\textup{Ablation Study}}
In this section, we delve into the ablation study of the proposed MoED and RAD mechanisms.
We perform experiments on nine medical image diagnosis datasets encompassing various medical modalities and tasks.
\textbf{Extended Data Table~\ref{tab:cls_ablation}} presents the experimental results.
For binary classification, accuracy is reported. 
Otherwise, the Macro-F1 score is reported.
Both MoED and RAD demonstrate consistent performance improvements for \ours{}, achieving average enhancements of 0.2030 and 0.2065, respectively. 
Compared with MoED, RAD exhibits superior performance, which can be attributed to its utilization of not only domain-specific knowledge from specialists but also information from the training database, thereby offering \ours{} more reliable references.
\par
To further assess the effectiveness and generalizability of RAD, we conducted experiments on broader types of downstream tasks.
In visual question answering and medical report generation tasks, developing a specialist model that consistently outperforms a GFM can be particularly challenging.
If specialist models underperform, they may fail to offer reliable guidance to the GFM, which could lead to a decline in overall performance.
Therefore, in the following experiments, we adopt the vision encoder of \ours{} as the retriever.
For medical image diagnosis tasks, following previous practice, we retrieve the labels of the five most similar cases. 
In visual question answering and medical report generation tasks, we retrieve the most similar images and then incorporate their corresponding annotations into the input.
Additionally, we also introduce Med-Flamingo~\cite{moor2023medflamingo} as a baseline model, which showcases impressive few-show learning ability.
\par
\textbf{Extended Data Table~\ref{tab:rag}} presents the results of medical image diagnosis, visual question answering, and medical report generation tasks.
The ``voting'' column reflects the outcomes derived from a voting mechanism based on the labels of the retrieved samples.
For medical report generation, we take the top-1 retrieved report as the ``voting'' results.
Notably, both Med-Flamingo and \ours{} exhibit significant and consistent improvements across most datasets and tasks, underscoring the generalizability of RAD.
However, Med-Flamingo's performance is upper-bounded by ``voting'', indicating its heavy reliance on retrieved results for final diagnoses without adequately considering the image content.
For instance, in the medical report generation task, Med-Flamingo often tends to directly rephrase or even replicate the retrieved samples with minimal modification.
In contrast, \ours{} consistently surpasses the ``Voting'' results across most downstream datasets, indicating that \ours{} not only considers the retrieved results but also leverages its intrinsic knowledge for diagnosing the test images.
Experiments validate the effectiveness of RAD, which can still enhance the capabilities of the GFM even in the absence of specialists, thereby boarding its application scenarios.

\section{Discussion}
To the best of our knowledge, we are the first to investigate the synergy between the GFM and specialist models.
Concretely, we propose a collaborative framework Generalist-Specialist Collaboration (GSCo), to leverage the generalist's in-context learning abilities alongside the specialists' domain-specific knowledge.
GSCo consists of two stages, namely the construction of GFM and specialists, and collaborative inference on downstream tasks.
In the construction stage, we first develop \ours{}, the largest open-source GFM tailored for medicine, capable of handling a wide range of medical tasks and modalities.
\ours{} also exhibits remarkable proficiency in both instruction-following and in-context learning, providing a solid foundation for cooperation with specialist models.
Meanwhile, a series of lightweight specialists are tailored for specific downstream tasks with low computational overhead.
In the collaborative inference stage, Mixture-of-Expert Diagnosis (MoED) and Retrieval-Augmented Diagnosis (RAD) are proposed as the core mechanisms of the cooperation.
MoED integrates predictions from specialists as reference diagnoses, while RAD employs these models to retrieve similar cases, collectively providing \ours{} with in-context information to facilitate medical image analysis.
To comprehensively evaluate \ours{} and GSCo, we curate the largest benchmark in medical GFM, which consists of 28 datasets and about 250K testing samples, encompassing diverse medical modalities and tasks.
Extensive qualitative and quantitative experiments highlight the following perspectives of our study.
\par
\textbf{\ours{} excels in understanding and analysis.}
Compared with previous methods~\cite{chen2023internvl, li2024llavamed, moor2023medflamingo, wu2023radfm, tu2024towards}, \ours{} exhibits significant advantages in two key aspects.
Firstly, as a generalist foundation model, \ours{} demonstrates remarkable generalizability in medical image analysis, enabling it to process a broader scope of medical modalities, including radiology, pathology, dermatology, ophthalmology, and gastroenterology, and achieving state-of-the-art performance across various tasks, such as visual question answering, medical report generation, and medical image diagnosis.
Secondly, \ours{} showcases exceptional capabilities in instruction following and in-context learning. 
This enhances the model's flexibility in handling different tasks and leveraging external knowledge, providing a solid foundation for effective collaboration with specialist models.
These advantages can be attributed to our meticulously curated training corpus and the larger scale of \ours{}.
Concretely, the training corpus incorporates over 2 million samples across five distinct task types, thereby broadening the scope of \ours{}.
Additionally, with 40 billion parameters, \ours{} surpasses previous models, endowing it with superior inherent capabilities.
In the future, we plan to incorporate more diverse training corpora as well as a larger foundation model.
\par
\textbf{Instruction following and in-context learning enables the Generalist-Specialist Collaboration.}
In most previous methods~\cite{chen2023internvl, li2024llavamed, moor2023medflamingo, wu2023radfm, tu2024towards, zhang2024generalist, yang2023medmnist}, both GFMs and specialist models independently handle the downstream tasks with their inherent capabilities.
GFMs are renowned for their generalizability and flexibility, enabling them to perform a variety of medical tasks across different modalities using a single model.
In contrast, specialist models are esteemed for their precision and efficiency, achieving satisfactory performance by tailoring them to specific downstream tasks with low computational consumption.
This work aims to explore the collaboration between generalist and specialist models, wherein instruction following and in-context learning, which previous methods overlooked, act as essential links that facilitate their integration.
Experimental results demonstrate that with exceptional instruction following and in-context learning capabilities, \ours{} can effectively collaborate with specialist models and achieve SOTA performance on downstream tasks.
\par
\textbf{GSCo presents a novel paradigm in the clinical application of GFM and specialists.}
In clinical practice, collaboration among healthcare professionals is not only common but also essential.
This study illustrates that through the synergistic collaboration between the GFM and specialist models, we can achieve superior performance on downstream tasks.
GSCo presents a new paradigm in the clinical application of both GFM and specialists.
Specifically, when confronted with out-of-domain tasks or data, rather than investing substantial resources to fine-tune the GFM, we can efficiently adapt lightweight specialist models with minimal resource expenditure.
Additionally, due to the stringent privacy regulations governing most medical data, including Protected Health Information (PHI), directly fine-tuning the GFM on data from multiple sources is often impractical.
Instead, under GSCo, we can integrate the knowledge from these private datasets by training the separate specialist models within their institutions independently, thereby safeguarding the confidentiality of medical data.
\par
\textbf{Mixture-of-expert diagnosis and retrieval-augmented diagnosis are effective collaboration strategies between the generalist and specialist models.}
In this study, we propose two mechanisms for providing \ours{} with the guidance of the specialists, namely mixture-of-expert diagnosis and retrieval-augmented diagnosis.
Mixture-of-expert diagnosis leverages the specialist model's predictions as references, while retrieval-augmented diagnosis treats the specialist model as a retriever to access relevant information in the training dataset.
In such collaboration, the roles of \ours{} and specialists are distinct.
The specialist model plays a pivotal role in equipping the GFM with reference guidance to enhance its generalization capabilities, particularly for out-of-domain tasks.
Meanwhile, \ours{} functions as a decision-maker with a wealth of medical knowledge.
Through extensive experiments, it has been demonstrated that these mechanisms enable the GFM to benefit from the specialist model's expertise, resulting in a robust and generalizable AI for medicine.
\par
\textbf{Limitations and further directions.}
Despite the advancements, the experimental results also reveal existing limitations, providing clear directions for future enhancements and research.
Firstly, while GSCo has demonstrated satisfactory performance across various public datasets, further exploration of its application in clinical practice is necessary to substantiate the superiority of the methodology. 
Secondly, multimodal retrieval strategies should be further explored. 
GSCo has showcased the significant potential of retrieval-augmented generation in medical image analysis.
However, the current vision-based retrieval method still struggles to ensure the accuracy of the retrieved samples.
Lastly, diverse collaborative paradigms require exploration. 
While GSCo has validated the effectiveness of collaboration, we anticipate observing more interaction patterns between generalists and specialists. 
For instance, incorporating Chain of Thought (CoT)~\cite{wei2022chain} could be a viable approach to facilitating collaboration between generalists and specialists.

\newpage
\section{Method}
\subsection*{Construction of GFM and Specialists}
\subsubsection*{Diagnosis-Guided Bootstrapping and Medical Image Description dataset}
\label{sec:dgb}
In this section, we introduce the proposed Diagnosis-Guided Bootstrapping (DGB) and Medical Image Description (DES) datasets.
To leverage the abundant medical image diagnosis datasets, we first present the DGB dataset.
Previous methods~\cite{li2024llavamed, wu2023radfm} have constructed instruction-tuning datasets primarily derived from PubMed research articles, to alleviate the scarcity of vision-language data in the medical domain.
Although these strategies have successfully assembled large-scale training corpora, they present two significant limitations.
Firstly, they rely solely on textual information, neglecting visual elements, which may lead to inconsistencies in descriptions.
Secondly, the content of research articles might lack reliability and accuracy, thereby introducing noise into the training corpus.
\par
\begin{figure}[!h]
\centering
\includegraphics[width=\linewidth, page=1]{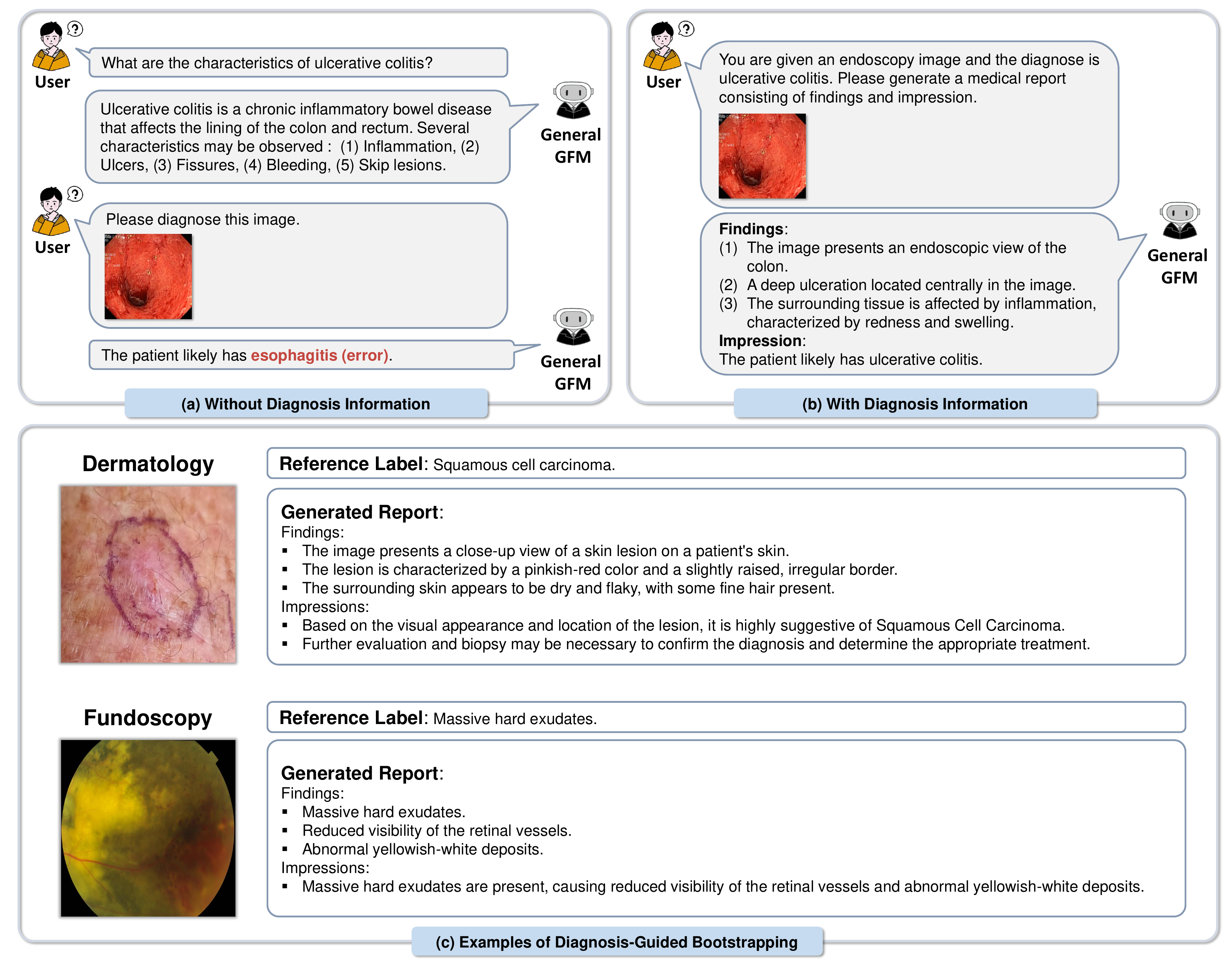}
\caption{
\textbf{(a)} 
The generalist foundation model in the general domain has the knowledge about diseases but it is challenging to correlate the knowledge with querying images, leading to \textbf{\textcolor{red1}{incorrect diagnosis}}.
\textbf{(b)} 
Guided by the correct diagnosis, the generalist foundation model generates a detailed medical report consisting of findings and impressions.
\textbf{(c)} Examples of the generated reports across different medical modalities.
}
\label{fig:dgb}
\end{figure}

In contrast, we propose to generate the instruction tuning dataset based on medical image diagnosis datasets, exploiting multi-modal information and human-verified annotations.
As shown in \textbf{Fig.~\ref{fig:dgb} (a)}, we observe that GFMs in the general domain~\cite{chen2023internvl, liu2024visual} exhibit a comprehensive understanding of disease-related information and associated symptoms, owing to extensive training on diverse corpora.
Nonetheless, these models encounter difficulties in correlating this knowledge with specific medical images, leading to inaccurate diagnostic predictions.
Conversely, when provided with specific disease and modality information alongside the given image, the GFM in the general domain is capable of generating high-quality medical reports with accurate diagnoses.
For example, \textbf{Fig.~\ref{fig:dgb} (b)} presents a generated report on ``ulcerative colitis'', where findings enumerate the observations in the image and the impression encapsulates the conclusion.
\par
Motivated by the above observations, we propose a diagnosis-guided bootstrapping strategy that leverages both visual and textual information to construct the instruction-tuning dataset. 
Specifically, we format the instruction as shown in \textbf{Fig.~\ref{fig:dgbprompt}}.
We provide the model with information about the modality and disease of the medical image and require it to generate a detailed report.
\begin{figure}[h]
\begin{promptenv}{Prompt for Diagnosis-Guided Bootstrapping}
You are a helpful medical assistant. \\
Your task is medical report generation. \\
You are given a \textbf{\{Modality\}} image, and the diagnosis is \textbf{\{Disease\}}. \\
You need to provide a medical report consisting of findings and impressions.
\end{promptenv}
\caption{
Prompt for diagnosis-guided bootstrapping.
\textbf{\{Modality\}} and \textbf{\{Disease\}} are placeholders for the corresponding information.}
\label{fig:dgbprompt}
\vspace{-10pt}
\end{figure}
\par
In contrast to previous works~\cite{li2024llavamed, wu2023radfm}, which generated data from textual information only, our approach offers two prominent advantages.
Firstly, it facilitates the utilization of numerous human-verified label-level annotated datasets in medicine.
Secondly, incorporating both visual and textual information ensures that the generated information remains pertinent to the accompanying images.
Following this method, we build the DGB dataset encompassing diverse medical image modalities. 
\par
Meanwhile, to augment the diversity of our training data, we present the Medical Image Description (DES) datasets.
Specifically, we collect the image-based case studies from OpenI~\cite{demner2012design} and then employ the GFM~\cite{chen2023internvl} to write a description of the image from a medical perspective, integrating both the image and associated textual information.
Additionally, we also exclude the details that are not inferable directly from the image, such as the patient's name or age, to maintain focus on diagnostically relevant visual features.
\subsubsection*{Medical Instruction Tuning}
To advance the multifaceted capabilities of the GFM, following previous methods~\cite{li2024llavamed, wu2023radfm, moor2023medflamingo}, we incorporate visual question answering~\cite{he2020pathvqa, lau2018visual, liu2021slake, zhang2023pmc, wu2023radfm} and medical report generation~\cite{johnson2019mimiccxr, demner2016iu} datasets into our training corpus.
Overall, as depicted in \textbf{Fig.~\ref{fig:main} (b)}, our training corpus encompasses five different types of items: medical image diagnosis (CLS), medical report generation (MRG), visual question answering (VQA), diagnosis-guided bootstrapping (DGB), and medical image description (DES). 
We meticulously crafted the prompt template for each type, as listed in \textbf{Extended Data Fig.~\ref{fig:sftprompt}}.
More detailed information about the training datasets is presented in \textbf{Section~\ref{sec:training_dataset_and_instruction_prompt}}.
The language modeling loss is utilized as the loss function to train the model.

\subsection*{Collaborative Inference on Downstream Tasks}
\subsubsection*{Mixture-of-Expert Diagnosis}
When addressing a specific downstream task, training a lightweight specialist model can often be more practical than utilizing GFMs. 
This advantage is primarily due to the significantly lower training overhead while still achieving satisfactory performance. 
In this study, we investigate the collaborative potential between generalist and specialist models, to enhance outcomes on downstream tasks.
\par
\begin{figure}[!t]
\centering
\includegraphics[width=\linewidth, page=1]{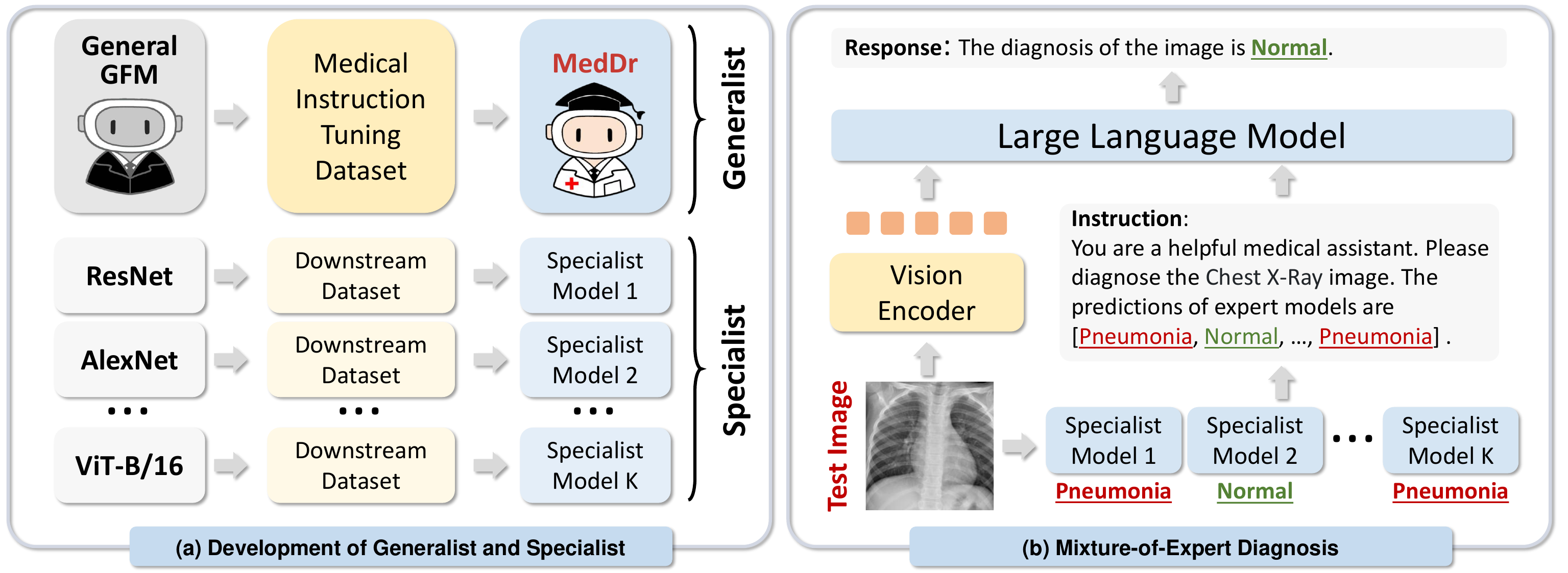}
\caption{
Mixture-of-Expert Diagnosis.
(a) \ours{} is a generalist foundation model trained on a large-scale medical instruction tuning dataset.
The specialist models are lightweight models trained on specific downstream datasets, requiring much lower training and inference consumption.
(b) During the inference on downstream tasks, the test image is fed into the specialist models first and their responses will act as the reference for \ours{}.
}
\label{fig:moed_method}
\end{figure}

As shown in \textbf{Fig.~\ref{fig:moed_method}}, we select several lightweight models and train them on the downstream dataset, designating these as specialist models.
Unlike the GFM, these specialist models are tailored to specific downstream tasks, yielding superior performance on these tasks due to expert knowledge.
During the inference phase, we first input the testing image into the specialist models and utilize their predictions as a guiding reference, which is subsequently incorporated into the instruction. 
\ours{} integrates both the testing image and the reference predictions to render the final diagnosis.
\begin{figure}[!t]
\centering
\includegraphics[width=\linewidth, page=1]{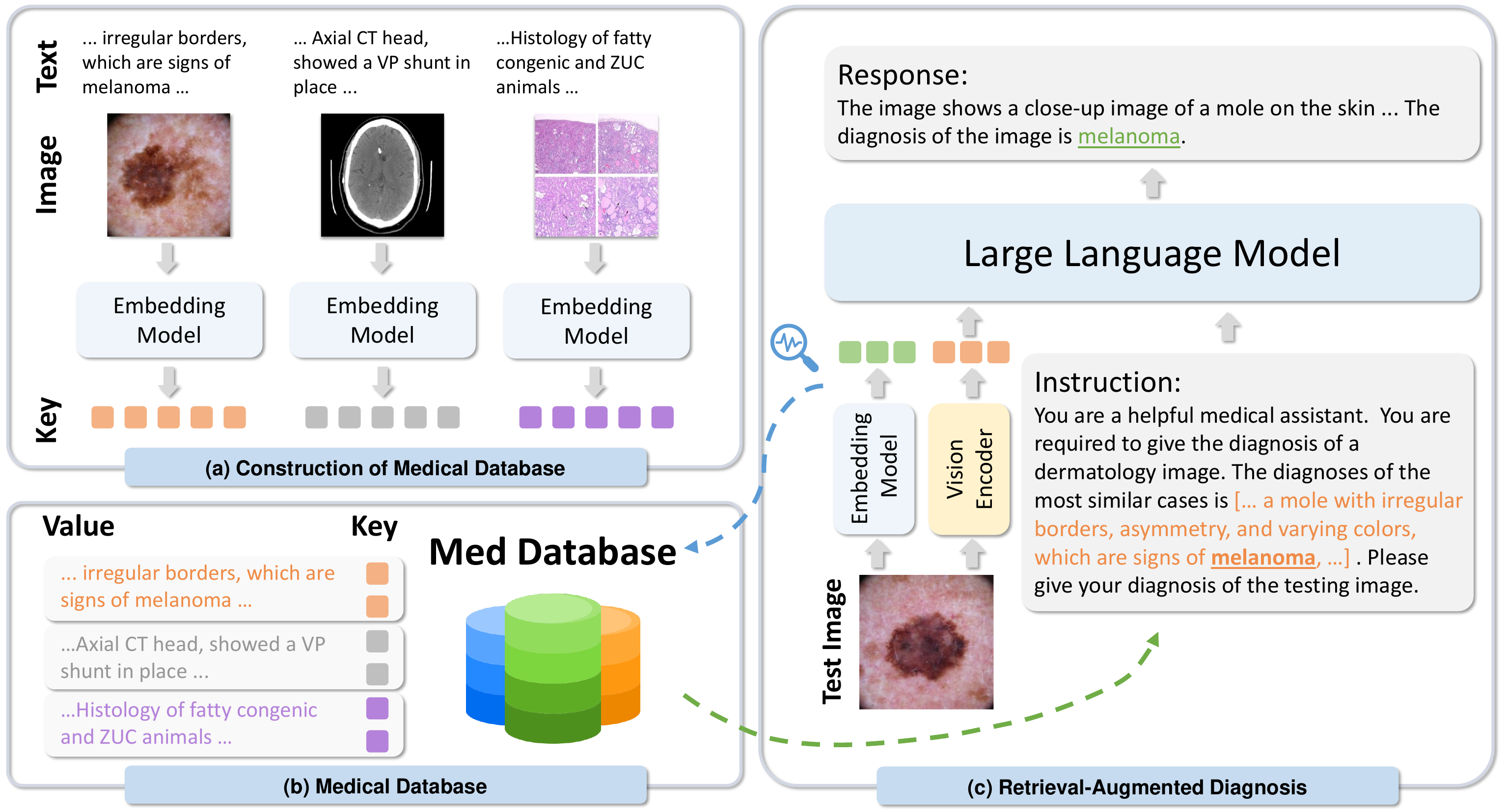}
\caption{
Retrieval-Augmented Diagnosis.
(a) An embedding model is employed to extract visual embeddings of images in medical image-text pairs.
(b) Each database entry comprises meta-information along with its indexed embedding.
(c) During the inference, we take the embedding of the test image to query the database and retrieve similar samples.
The information from the retrieved samples is integrated into the instruction as contextual information.
Both the test image and the instruction are then fed into \ours{}.
}
\label{fig:rad_method}
\end{figure}
\subsubsection*{Retrieval-Augmented Diagnosis}
GFMs have demonstrated significant capabilities but still face challenges when encountering out-of-domain data~\cite{chen2023internvl, van2023radadapt}.
To exploit the training corpus exhaustively and enhance the capability on out-of-domain datasets, we propose a collaborative mechanism, Retrieval-Augmented Diagnosis.
\par
\textbf{Fig.~\ref{fig:rad_method}} illustrates the proposed method, where the database is built on the training data across multiple medical tasks and modalities.
Concretely, to build the database, given an image-text pair, we encode the image by an embedding model and take this visual embedding as the key while the text is the value.
During the inference time, we first employ the embedding model to encode the test image.
Then, the visual embedding of the test image is used to query the database by calculating the cosine similarity between the query and keys.
The top-$k$ most similar items from the database are retrieved, and then their meta information is incorporated into the instruction as additional textual clues to help the model make medical decisions.
\par
In this study, we first explore utilizing the specialist model as the retriever. 
After fine-tuning on the specific downstream dataset, this specialist model is capable of generating more discriminative embeddings, thereby enhancing retrieval accuracy.
Furthermore, we also explore the application of the vision encoder from \ours{} as the retriever, which showcases two advantages.
First, it effectively eliminates the need for an additional embedding module, which in turn reduces associated computational costs. By leveraging the vision encoder from \ours{} as the embedding module, we can directly use the intermediate results as embeddings for queries during inference without incurring any extra overhead.
Second, this approach is applicable to a wider array of scenarios, particularly in situations where acquiring a qualified specialist model poses challenges.
\subsection*{Implementation Details}
In this work, we employ InternVL~\cite{chen2023internvl}, a state-of-the-art GFM in the general domain, as our foundation model, which contains about 40B parameters, consisting of a 6B vision encoder and a 34B large language model.
The input image is resized to 448$\times$448 pixels.
The model is fine-tuned on both collected and generated data. 
The number of training samples is about 2M.
Please refer to \textbf{Section \ref{sec:training_dataset_and_instruction_prompt}} for detailed information about the training dataset and instruction prompt.
The instruction tuning recipe follows the suggestions provided by InternVL.
We fix all parameters except for the LoRA component, which is composed of approximately 0.1B parameters, accounting for 0.4\% of the total parameters.
Meanwhile, we also leverage DeepSpeed ZeRO Stage 3~\cite{rajbhandari2021zero} to optimize the training procedure.
The model is trained on 16 NVIDIA H800 GPUs for two epochs within 72 hours.
The specialist models are trained on a single NVIDIA 4090 GPU and the corresponding recipe is presented in \textbf{Extended Data Table~\ref{tab:spehyper}}.
\subsection*{State-of-the-art Generalist and Specialist Models}
\label{sec:gen_spe_model}
In this study, we select 4 open-source models in both the general and medical domains as the baseline GFMs.
\begin{itemize}
\item
\textbf{RadFM}~\cite{wu2023radfm} mainly focuses on the radiology modality. 
It consists of a 3D ViT as the vision backbone and PMC-LLaMA-13B~\cite{wu2023pmcllama} as the LLM.
The model is first trained on 16M noisy pre-training data and then fine-tuned on 3M in-domain data.
\item
\textbf{LLaVA-Med}~\cite{li2024llavamed} is built on the pre-trained LLaVA~\cite{liu2024visual}. 
It is fine-tuned on about 600K concept alignment samples and 60K instruction tuning samples in one day with 8 A100 GPUs.
\item
\textbf{Med-Flamingo}~\cite{moor2023medflamingo} is developed based on OpenFlamingo-9B~\cite{awadalla2023openflamingo}, which can handles multiple images interleaved with texts.
It is trained on large-scale interleaved datasets based on medical textbooks and the PMC-OA dataset~\cite{lin2023pmc}.
\item 
\textbf{InternVL}~\cite{chen2023internvl} is one of the most powerful open-source large-scale vision-language models in the general domain. 
It obtains SOTA performance on multi-modal tasks in the general domain.
\end{itemize}
For these GFMs, we reproduce the above models based on their open-source checkpoint and evaluate the model using the same test data. 
The test prompt is set up according to their official implementations (\textbf{Extended Data Table~\ref{tab:public_code}}).
\par
For the choice of the specialist models, following~\cite{doerrich2024rethinking}, we select ten representative computer vision models and train them on the specific downstream datasets:
\begin{itemize}
\item \textbf{VGG16}~\cite{simonyan2014very} is a convolution-based model. It consists of 16 layers and is known for its simplicity and effectiveness. \par
\item \textbf{AlexNet}~\cite{krizhevsky2012imagenet} is a convolution-based model. It won the ImageNet Large Scale Visual Recognition Challenge (ILSVRC) in 2012 and popularized the use of convolutional neural networks (CNNs). \par
\item \textbf{ResNet-18}~\cite{he2016deep} is a convolution-based model. It introduced residual connections that allow gradients to flow directly through the network, enabling the training of very deep networks. \par
\item \textbf{DenseNet-121}~\cite{huang2017densely} is a convolution-based model. It encourages feature reuse by connecting each layer to every other layer in a feed-forward fashion.\par
\item \textbf{EfficientNet-B4}~\cite{tan2019efficientnet} is a convolution-based model. It balances depth, width, and resolution to achieve high performance with fewer parameters. It is efficient in terms of both accuracy and computation.\par
\item \textbf{ViT-B/16}~\cite{dosovitskiy2020image} is a Transformer-based model. It applies the Transformer architecture to image data, achieving competitive results on various vision tasks. \par
\item \textbf{CLIP ViT-B/16}~\cite{radford2021learning} is a Transformer-based model. It learns to associate images and text in a joint embedding space, enabling zero-shot classification and other multimodal tasks.\par
\item \textbf{EVA-02 ViT-B/16}~\cite{fang2023eva} is a Transformer-based model. It improves the training techniques for CLIP at scale and achieves superior performance with significantly smaller training costs. \par
\item \textbf{DINO ViT-B/16}~\cite{caron2021emerging} is a Transformer-based model. It leverages clustering and momentum encoders and achieves strong performance without using labeled data.\par
\item \textbf{SAM ViT-B/16}~\cite{kirillov2023segment} is a Transformer-based model, which is designed to be a general-purpose model for image segmentation, capable of segmenting any object in an image with minimal user input.\par
\end{itemize}
The details of these specialist models are listed in \textbf{Extended Data Table~\ref{tab:spemodel}}.
The fine-tuning recipe on the downstream dataset is shown in \textbf{Extended Data Table~\ref{tab:spehyper}}.
Compared with the GFMs, these models have much fewer parameters and can be trained on consumer-level hardware, like NVIDIA 4090 GPU.
\subsection*{Training Dataset}
\label{sec:training_dataset_and_instruction_prompt}
In this section, we introduce the training dataset in detail and \textbf{Extended Data Fig.~\ref{fig:sftprompt}} presents the prompt template for each type of dataset.

\subsubsection*{Visual Question Answering Datasets}
\begin{itemize}
\item
\textbf{SLAKE}~\cite{liu2021slake} is a bilingual radiology VQA dataset comprising 642 images and 14K questions. 
We only use the English part of the training split, which consists of 4,919 question-answer pairs.
\item
\textbf{VQA-RAD}~\cite{lau2018visual} is a manually constructed dataset where clinicians asked naturally occurring questions of radiology images and provided reference answers.
Following the official split, we use 3,064 question-answer pairs of the training set.
\item
\textbf{Path-VQA}~\cite{he2020pathvqa} consists of 32,799 open-ended questions from 4,998 pathology images, where each question is manually checked to ensure correctness.
Following the official split, we use 19,755 question-answer pairs of the training set.
\item
\textbf{PMC-VQA}~\cite{zhang2023pmc} is a large-scale medical visual question-answering dataset built from image-text pairs from PubMed Central, covering broader medical image modalities.
Following the official split, we use 152,603 question-answer pairs of the training set.
\item
\textbf{PMC-CaseReport}~\cite{wu2023radfm} is an auto-generated visual question-answering dataset based on the case report papers in the PMC-Inline dataset.
Following the official split, we use 254,105 question-answer pairs of the training set. 
\end{itemize}

\subsubsection*{Medical Report Generation Datasets}
\begin{itemize}
\item
\textbf{MIMIC-CXR}~\cite{johnson2019mimiccxr} presents 371,920 chest X-rays associated with 227,943 imaging studies from 65,079 patients.
Following RadFM~\cite{wu2023radfm} and R2Gen~\cite{chen-emnlp-2020-r2gen}, we use 337,292 cases for training.
\item
\textbf{IU-Xray}~\cite{demner2016iu} is a set of chest X-ray images paired with their corresponding diagnostic reports. The dataset contains 7,470 pairs of images and reports.
Following R2Gen~\cite{chen-emnlp-2020-r2gen}, we use 4,730 cases from the training split.
\end{itemize}

\subsubsection*{Medical Image Diagnosis Datasets}
\begin{itemize}
\item
\textbf{VinDr-SpineXR}~\cite{nguyen2021vindr} is a large annotated medical image dataset for spinal lesion detection and classification from radiographs.
Following RadFM~\cite{chen-emnlp-2020-r2gen}, we use 6,129 samples for training.
\item
\textbf{VinDr-PCXR}~\cite{pham2022vindr} is an open-source large-scale pediatric chest X-ray dataset for the interpretation of common thoracic diseases.
Following RadFM~\cite{chen-emnlp-2020-r2gen}, we use 4,585 samples for training.
\item
\textbf{VinDr-Mammo}~\cite{nguyen2023vindr} is a large-scale benchmark dataset for computer-aided detection and diagnosis in full-field digital mammography.
Following RadFM~\cite{chen-emnlp-2020-r2gen}, we use 6,047 samples for training.
\item
\textbf{VinDr-CXR}~\cite{nguyen2020vindrcxr} is an open large-scale dataset of chest X-rays with radiologist’s annotations. The training set contains 15,000 scans, and 3 radiologists independently label each image.
Following the official split, we use 45,000 samples for training.
\item
\textbf{CheXpert}~\cite{irvin2019chexpert} is a large public dataset for chest radiograph interpretation, consisting of 224,316 chest radiographs of 65,240 patients.
Following the official split, we use 223,414 samples for training.
\item
\textbf{ChestX-ray14}~\cite{wang2017chestx} is a medical imaging dataset which comprises 112,120 frontal-view X-ray images of 30,805 patients with the text-mined fourteen common disease labels.
Following the official split, we use 86,524 samples for training.
\item
\textbf{PCam200}~\cite{kawai2023large} is a public pathological H\&E image dataset made in the same manner from Camelyon2016 challenge dataset~\cite{bejnordi2017diagnostic}.
Following the official split, we use 28,539 samples for training.
\item
\textbf{PAD-UFES-20}~\cite{pacheco2020pad} is a dermatology classification dataset consisting of 2,298 images for six different diagnostics.
We use all of the 2,298 samples for training.
\item
\textbf{DermNet}~\cite{kaggleDermnet}  consists of dermatology images of 23 types of skin diseases taken from DermNet.
Following the official split, we use 15,557 samples for training.
\item
\textbf{HAM10000}~\cite{tschandl2018ham10000} is a large collection of multi-source dermatoscopic images of pigmented lesions.
Following the official split, we use 10,015 samples for training.
\item
\textbf{ISIC2020}~\cite{rotemberg2021patient} is a dataset of the SIIM-ISIC Melanoma Classification Challenge 2020.
The dataset contains 33,126 dermoscopic training images of unique benign and malignant skin lesions from over 2,000 patients.
Following the official split, we use 33,126 samples for training.
\item
\textbf{Kvasir}~\cite{pogorelov2017kvasir} is a multi-class image dataset for computer-aided gastrointestinal disease detection.
Following the official split, we use 8,000 samples for training.
\item
\textbf{Kvasir Capsule}~\cite{Smedsrud2021} is an endoscopy dataset consisting of 47,238 images with annotations of anatomical landmarks and pathological and normal findings.
Following the official split, we use 47,238 samples for training.
\item
\textbf{WCE}~\cite{kaggleWCE} is a curated colon disease dataset based on Kvasir~\cite{pogorelov2017kvasir} and ETIS-Larib-Polyp DB~\cite{silva2014toward}.
Following the official split, we use 3,200 samples for training.
\item
\textbf{GastroVision}~\cite{jha2023gastrovision} is a multi-center open-access gastrointestinal (GI) endoscopy dataset that includes different anatomical landmarks, pathological abnormalities, polyp removal cases, and normal findings from the GI tract.
We use all of the 8,000 samples for training.
\item
\textbf{ODIR}~\cite{li2021benchmark} is a structured ophthalmic database of 5,000 patients with age, color fundus photographs from left and right eyes, and doctors' diagnostic keywords.
Following the official split, we use 6,392 samples for training.
\item
\textbf{Fundus1000}~\cite{cen2021automatic} contains 1,000 fundus images with 39 categories.
We use all of the 1,000 samples for training.
\item
\textbf{RFMiD2.0}~\cite{data8020029} is a multi-label dataset including around 860 retinal fundus images annotated by three eye specialists.
Following the official split, we use 455 samples for training.
\item
\textbf{Retinal OCT-C8}~\cite{9740985} is a large-scale dataset for ophthalmic research containing 24,000 optical coherence tomography (OCT) images that are organized into eight categories.
Following the official split, we use 18,000 samples for training.
\item
\textbf{UltraBreast} is a private breast ultrasound dataset that contains 45,896 cases that are labeled benign or malignant.
\end{itemize}

\subsubsection*{Synthetic Datasets}
\noindent\textbf{Diagnosis-Guided Bootstrapping Dataset.} As introduced in Section~\ref{sec:dgb}, we constructed a large-scale medical report dataset across diverse medical modalities. 
Concretely, we generate 196,760 samples in total based on the VinDr-SpineXR~\cite{nguyen2021vindr}, VinDr-PCXR~\cite{pham2022vindr}, VinDr-Mammo~\cite{nguyen2023vindr}, VinDr-CXR~\cite{nguyen2020vindrcxr}, ChestX-ray14~\cite{wang2017chestx}, PAD-UFES-20~\cite{pacheco2020pad}, Dermnet~\cite{kaggleDermnet}, Kvasir~\cite{pogorelov2017kvasir}, WCE~\cite{kaggleWCE}, Kvasir Capsule~\cite{Smedsrud2021}, ODIR~\cite{li2021benchmark}, Fundus1000~\cite{cen2021automatic} and RFMiD2.0~\cite{data8020029} datasets.
\par
\noindent\textbf{Medical Image Description Dataset.}
To augment the diversity of our training data, we also collected 245,371 image-based case studies from OpenI~\cite{demner2012design}.
We summarize the title of the case and the image caption based on the image by InternVL~\cite{chen2023internvl} and obtain high-quality images and corresponding text summaries.

\subsection*{Out-of-Domain Benchmark Dataset}
\subsubsection*{Visual Question Answering Datasets}
\begin{itemize}
\item 
\textbf{VQA-Med}~\cite{ImageCLEFVQA-Med2019} focuses on radiology images and consists of four main categories of questions: modality, plane, organ system, and abnormality.
Following the official split, we use 500 items for testing.
\item 
\textbf{OmniMedVQA}~\cite{hu2024omnimedvqa} is a large-scale comprehensive evaluation benchmark dataset for the medical GFMs.
Due to the overlap with our training data, we exclude some data to prevent data leakage and only use the Disease Diagnosis subset.
The total number of the testing samples is 51,977.
\end{itemize}
\subsubsection*{Medical Image Diagnosis Datasets}
\begin{itemize}
\item 
\textbf{PneumoniaMNIST}~\cite{yang2023medmnist} is a binary classification dataset about chest X-ray.
Following the official split, we use 624 items for testing.
\item 
\textbf{BreastMNIST}~\cite{yang2023medmnist} is a binary classification dataset of breast ultrasound.
Following the official split, we use 156 samples for testing.
\item 
\textbf{OrganAMNIST}~\cite{yang2023medmnist} is a multi-class classification dataset of abdominal CT.
Following the official split, we use 17,778 items for testing.
\item 
\textbf{PathMNIST}~\cite{yang2023medmnist} is a multi-class classification dataset of colon pathology.
Following the official split, we use 7,180 items for testing.
\item 
\textbf{OCTMNIST}~\cite{yang2023medmnist} a multi-class classification dataset of retinal OCT.
Following the official split, we use 1,000 items for testing.
\item 
\textbf{ChestMNIST}~\cite{yang2023medmnist} is a multi-label dataset of chest X-ray.
Following the official split, we use 22,433 samples for testing.
\item 
\textbf{CBIS-DDSM}~\cite{sawyer2016curated} contains images for screening mammography.
The original dataset contains images of cases with three conditions of breast cancer: BENIGN, BENIGN\_WITHOUT\_CALLBACK, and MALIGNANT.
Due to the insufficient information in the text to discriminate between BENIGN and BENIGN\_WITHOUT\_CALLBACK, we formulate it as a binary classification task.
Following the official split, we use 326 samples from the CALC subset and 378 samples from MASS for testing.
\item 
\textbf{FMC-Colon}~\cite{wang2023real} is a pathological tumor tissue classification dataset and requests the model to determine whether the sample is positive or negative.
Following the official split, we use 4,355 samples for testing.
\item 
\textbf{FMC-Endo}~\cite{wang2023real} is a colonoscopy lesion classification dataset and consists of four different lesion types.
Following the official split, we use 2,055 samples for testing.
\item 
\textbf{FMC-Chest}~\cite{wang2023real} is a thoracic disease screening dataset and covers 19 common thoracic abnormalities.
Following the official split, we use 2,708 samples for testing.
\item 
\textbf{Derm7pt}~\cite{kawahara2018seven} is a dataset for evaluating computerized image-based prediction of the 7-point skin lesion malignancy checklist. 
Following the official split, we use 395 samples for testing.
\item 
\textbf{BRSET}~\cite{nakayama2023brazilian} is a multi-labeled ophthalmological dataset designed to improve scientific community development and validate machine learning models.
We randomly divided the dataset into training and testing splits with an 8:2 ratio.
The number of testing samples is 3,254.
\end{itemize}
\subsection*{Evaluation Metrics} 
\subsubsection*{Medical Image Diagnosis}
For the medical image diagnosis datasets, accuracy and F1-Score are exploited for evaluation. 
Accuracy is calculated as
$$
\text { Accuracy }=\frac{1}{N} \sum_i^N 1\left(y_i=\hat{y}_i\right),
$$
where $y$ is a tensor of target values, and $\hat{y}$ is a tensor of predictions.
F1-Score is defined based on recall and precision as follows,
$$
\text { Recall }=\frac{\mathrm{TP}}{\mathrm{TP}+\mathrm{FN}},
$$
$$
\text { Precision }=\frac{\mathrm{TP}}{\mathrm{TP}+\mathrm{FP}},
$$
$$
\text { F1-Score }=2 \cdot \frac{\text { Precision } * \text { Recall }}{\text { Precision }+ \text { Recall }},
$$
where TP and FP represent the number of true positives and false positives respectively.
Especially, for multi-class and multi-label classification datasets, we calculate both Macro-F1 Score and Micro-F1 Score.
\subsubsection*{Visual Question Answering}
For the dataset consisting of multiple choice questions~\cite{hu2024omnimedvqa}, we calculate the accuracy. 
For other visual question answering datasets, following MultiMedEval~\cite{royer2024multimedeval}, we first tokenize both prediction and answer and compute precision and recall.
For close-ended questions, the prediction is correct if its recall is at least 0.5.
For open-ended questions, the prediction is correct if its recall is at least 0.75.
We report the accuracy of close-ended questions and the accuracy and recall of the open-ended questions.
The overall F1-Score and recall are also reported.
Additionally, following~\cite{wu2023radfm}, we compute the BLEU-1 score as follows:
$$
\text{BLEU-1}=\mathrm{BP} \cdot \exp \left(\sum_{n=1}^N w_n \cdot \log p_n\right),
$$
where \( \text{BP} \) is the brevity penalty, \( p_n \) is the precision for n-grams, and \( w_n \) is the weight for n-gram precision. If the predicting result's length $c$ is greater than the reference length  $r$, then $\text{BP} = 1$.
If $c \leq r$, then $\text{BP} = \exp(1 - r/c)$.
This ensures that a shorter predicting result is penalized to prevent the system from favoring overly concise output. Since there is only one type of n-gram, $w_1=1$. The brevity penalty (BP) is used to adjust the BLEU score based on the length of the candidate translation compared to the reference translation. It's calculated as follows:

\subsubsection*{Medical Report Generation}
\label{sec:metric}
For medical report generation tasks, we utilize common n-gram-based metrics such as BLEU-1, BLEU-4, ROUGE-1, ROUGE-L, and METEOR~\cite{banerjee2005meteor}.
Here, ROUGE-1 is defined as follows,

\[
\text{ROUGE-1} = \frac{|\text{Recall} \cap \text{Reference}|}{|\text{Reference}|},
\]
where $|\text{Recall} \cap \text{Reference}|$ is the number of overlapping unigrams between the generated report and the reference report, whereas $|\text{Reference}|$ refers to the total number of unigrams in the reference text.
ROUGR-L is defined as
\[
\text{ROUGE-L} = \frac{F_{LCS}}{|\text{Reference}|},
\]
where \( F_{LCS} \) represents the F1 score of the longest common subsequence. 

The METEOR score is computed as:

\[
\text{METEOR} = \frac{1}{m} \cdot \sum_{g \in \text{gold}} \max_{h \in \text{hyp}} \text{Precision}(g, h),
\]
where $m$ is the number of gold standard (reference) sentences, and $\text{Precision}(g,h)$ refers to the precision score between a specific gold standard sentence ($g$) and a hypothesis sentence ($h$) from the set of all gold standard sentences ($\text{gold}$) and the set of all hypothesis sentences ($\text{hyp}$).
 
Moreover, we evaluate F1-RadGraph, which measures the F1 score between entities extracted from the reference and generated reports using RadGraph~\cite{jain1radgraph}:

\[
\text{F1-RadGraph} = 2 \cdot \frac{\text{Precision} \cdot \text{Recall}}{\text{Precision} + \text{Recall}}.
\]

We also compute CheXbert vector similarity~\cite{yu2023evaluating} using cosine similarity between the embedded reference and generated reports:

\[
\text{Cosine Similarity} = \frac{A \cdot B}{\|A\| \|B\|},
\]
where \( A \) and \( B \) are the vectors of the reference and generated reports, respectively.

\backmatter
\subsection*{Data availability}
The datasets used for building the training dataset are listed in \textbf{Extended Data Table~\ref{data_available_training}}. and the evaluation benchmark datasets are listed in \textbf{Extended Data Table~\ref{data_available_testing}}.

\subsection*{Code availability}
The implementation of \ours{} and GSCo framework will be available at \href{https://github.com/sunanhe/MedDr}{https://github.com/sunanhe/MedDr}.
The weights of \ours{} and specialist models will be avaliable in \href{https://huggingface.co/Sunanhe/MedDr_0401}{https://huggingface.co/Sunanhe/MedDr\_0401}.
The other public codes used in this study are listed in \textbf{Extended Data Table~\ref{tab:public_code}}.
\subsection*{Author contribution}
S.H., Y.N., and H.C. conceived and designed the work.
S.H., Y.N., contributed to the technical implementation and conducted experiments.
S.H., Y.N., Zhixuan.C., Zhiyuan.C., H.W., S.Y., Y.W., and Y.X. contributed to the data acquisition and organization.
L.L., H.X., X.L., M.W, Y.P, G.S, Z.X, X.W, Q.W., R.C.K.C, V.V, W.C.W.C., Y.Z, P.R., and K.Z. provided suggestions on the framework and experiments.
All authors contributed to the drafting and revising of the manuscript.
H.C. supervised the research.
\subsection*{Declarations}
The authors have no conflicts of interest to declare.
\subsection*{Ethics declarations}
This study has been reviewed and approved by the Human and Artefacts Research Ethics Committee (HAREC). The protocol number is HREP-2024-0212.
\subsection*{Acknowledgements}
This work was supported by the Hong Kong Innovation and Technology Commission (Project No. MHP/002/22 and TCPD/17-9), HKUST 30 for 30 Research Initiative Scheme, Cornell–HKUST Global Strategic Collaboration Awards, Asian Young Scientist Fellowship and the Research Grants Council of the Hong Kong (Project Reference Number: T45-401/22-N). 
\bigskip
\newpage
\bibliography{sn-bibliography}
\newpage
\begin{appendices}
\section{Extended Data}
\begin{sidewaystable*}[h]
\centering
\caption{
Detailed results on binary classification datasets in medical image diagnosis task.
Results are reported in terms of accuracy.
\textbf{\textcolor{red}{Red}} indicates the overall best results, and \textbf{\textcolor{blue}{Blue}} indicates the overall second-best results.
\textbf{Black} denotes the best results of each model group.
Numbers in parentheses indicate a 95\% confidence interval (CI).
}
\adjustbox{totalheight=0.6\textheight, width=1\linewidth}{
\begin{tabular}{l|p{3.8cm}<{\centering}|p{3.4cm}<{\centering}|p{3.4cm}<{\centering}|p{3.6cm}<{\centering}|p{3.8cm}<{\centering}|p{3.8cm}<{\centering}|p{2cm}<{\centering}}
\toprule
\rowcolor{lightgray}
\textbf{Method} & \textbf{PCam200} & \textbf{FMC-Colon} & \textbf{PneumoniaMNIST} & \textbf{BreastMNIST} & \textbf{CBIS-DDSM(MASS)} & \textbf{CBIS-DDSM(CALC)} & \textbf{Average}\\
\midrule
\rowcolor{lightgray!50}
\multicolumn{8}{c}{\textbf{Generalist Foundation Model} } \\
\multirow{2}{*}{RadFM} & 0.4423 & 0.3851 & 0.5865 & 0.3526 & 0.3889 & 0.4141 &\multirow{2}{*}{0.4283} \\
 & (0.4387,0.4459) & (0.3779,0.3931) & (0.5669,0.6046) & (0.3138,0.3919) & (0.3651,0.4126) & (0.3871,0.4409) &\\
 \hline
\multirow{2}{*}{LLaVA-Med} & 0.4986 & 0.4953 & 0.5000 & 0.3782 & \textbf{0.6111} & 0.6012 &\multirow{2}{*}{0.5141} \\
 & (0.4951,0.5022) & (0.4080,0.4271) & (0.4807,0.5192) & (0.3408,0.4148) & (0.5868,0.6353) & (0.5734,0.6275) &\\
 \hline
\multirow{2}{*}{Med-Flamingo} & 0.3986 & 0.3738 & 0.7131 & 0.3077 & \textbf{0.6111} & \textbf{0.6043} &\multirow{2}{*}{0.5014} \\
 & (0.3950,0.4025) & (0.3665,0.3810) & (0.6957,0.7309) & (0.2722,0.3438) & (0.5877,0.6349) & (0.5790,0.6304) &\\
 \hline
\multirow{2}{*}{InternVL} & 0.4986 & 0.4533 & 0.4583 & 0.4615 & 0.3889 & 0.3957 &\multirow{2}{*}{0.4427} \\
 & (0.4949,0.5025) & (0.4461,0.4610) & (0.4395,0.4763) & (0.4221,0.5032) & (0.3641,0.4125) & (0.3660,0.4211) &\\
 \hline
\multirow{2}{*}{MedDr} & \textbf{0.9089} & \textbf{0.6262} & \textbf{0.8734} & \textbf{0.7179} & 0.5132 & 0.5184 &\multirow{2}{*}{\textbf{0.6930}} \\
 & (0.9066,0.9110) & (0.6184,0.6334) & (0.8606,0.8862) & (0.6835,0.7541) & (0.4871,0.5388) & (0.4926,0.5450) &\\
\midrule
\rowcolor{lightgray!50}
\multicolumn{8}{c}{\textbf{Specialist Model} } \\
\multirow{2}{*}{VGG16} & \textbf{\textcolor{blue}{0.9203}} & \textbf{0.9621} & 0.8478 & 0.8782 & 0.6614 & \textbf{\textcolor{red}{0.6963}} &\multirow{2}{*}{0.8277} \\
 & (0.9183,0.9223) & (0.9593,0.9650) & (0.8328,0.8613) & (0.8545,0.9016) & (0.6378,0.6863) & (0.6721,0.7213) \\
 \hline
\multirow{2}{*}{AlexNet} & 0.8340 & 0.9093 & 0.8317 & 0.8782 & 0.6402 & 0.6871 &\multirow{2}{*}{0.7968} \\
 & (0.8314,0.8367) & (0.9049,0.9137) & (0.8170,0.8453) & (0.8497,0.9020) & (0.6166,0.6639) & (0.6617,0.7117) \\
 \hline
\multirow{2}{*}{ResNet-18} & 0.8816 & 0.9571 & 0.8910 & 0.8590 & 0.6243 & 0.5798 &\multirow{2}{*}{0.7988} \\
 & (0.8794,0.8839) & (0.9541,0.9599) & (0.8789,0.9026) & (0.8322,0.8852) & (0.6021,0.6473) & (0.5536,0.6062) \\
 \hline
\multirow{2}{*}{DenseNet-121} & 0.9005 & 0.9575 & 0.9022 & 0.8590 & \textbf{\textcolor{red}{0.7011}} & 0.6534 &\multirow{2}{*}{\textbf{0.8290}} \\
 & (0.8985,0.9026) & (0.9544,0.9606) & (0.8915,0.9137) & (0.8314,0.8849) & (0.6795,0.7229) & (0.6275,0.6802) \\
 \hline
\multirow{2}{*}{EfficientNet} & 0.8320 & 0.8994 & 0.8782 & 0.7500 & 0.5556 & 0.6135 &\multirow{2}{*}{0.7548} \\
 & (0.8293,0.8348) & (0.8949,0.9036) & (0.8640,0.8911) & (0.7160,0.7835) & (0.5309,0.5808) & (0.5844,0.6392) \\
 \hline
\multirow{2}{*}{ViT} & 0.9135 & 0.9240 & 0.8462 & \textbf{\textcolor{blue}{0.8974}} & 0.6693 & 0.6012 &\multirow{2}{*}{0.8086} \\
 & (0.9114,0.9156) & (0.9202,0.9279) & (0.8317,0.8602) & (0.8738,0.9195) & (0.6457,0.6920) & (0.5759,0.6277) \\
 \hline
\multirow{2}{*}{CLIP ViT} & 0.8603 & 0.9054 & 0.8478 & 0.7372 & 0.6376 & 0.5920 &\multirow{2}{*}{0.7634} \\
 & (0.8579,0.8627) & (0.9010,0.9096) & (0.8337,0.8621) & (0.7048,0.7727) & (0.6151,0.6613) & (0.5628,0.6182) \\
 \hline
\multirow{2}{*}{EVA-02 ViT} & 0.8836 & 0.9543 & \textbf{\textcolor{blue}{0.9038}} & 0.8590 & 0.6005 & 0.6166 &\multirow{2}{*}{0.8030} \\
 & (0.8811,0.8859) & (0.9510,0.9573) & (0.8931,0.9155) & (0.8307,0.8839) & (0.5745,0.6250) & (0.5908,0.6417) \\
 \hline
\multirow{2}{*}{DINO ViT} & 0.8811 & \textbf{0.9621} & 0.8141 & 0.8205 & 0.6667 & 0.5982 &\multirow{2}{*}{0.7905} \\
 & (0.8787,0.8834) & (0.9592,0.9649) & (0.8000,0.8286) & (0.7873,0.8497) & (0.6435,0.6897) & (0.5710,0.6248) \\
 \hline
\multirow{2}{*}{SAM ViT} & 0.9063 & 0.9451 & 0.8814 & 0.8269 & 0.6190 & 0.6564 &\multirow{2}{*}{0.8059} \\
 & (0.9039,0.9084) & (0.9416,0.9485) & (0.8689,0.8937) & (0.7968,0.8553) & (0.5934,0.6434) & (0.6307,0.6822) \\
\midrule
\rowcolor{lightgray!50}
\multicolumn{8}{c}{\textbf{Collaborative Method} } \\
\multirow{2}{*}{Voting} & 0.9132 & \textbf{\textcolor{red}{0.9711}} & 0.8702 & 0.8910 & \textbf{\textcolor{blue}{0.6958}} & 0.6564 &\multirow{2}{*}{\textbf{\textcolor{blue}{0.8330}}} \\
 & (0.9110,0.9153) & (0.9683,0.9733) & (0.8558,0.8826) & (0.8667,0.9130) & (0.6728,0.7168) & (0.6316,0.6822) \\
 \hline
\multirow{2}{*}{GSCo} & \textbf{\textcolor{red}{0.9321}} & \textbf{\textcolor{blue}{0.9699}} & \textbf{\textcolor{red}{0.9599}} & \textbf{\textcolor{red}{0.9295}} & 0.6878 & \textbf{\textcolor{blue}{0.6933}} &\multirow{2}{*}{\textbf{\textcolor{red}{0.8621}}} \\
 & (0.9302,0.9340) & (0.9673,0.9724) & (0.9518,0.9670) & (0.9091,0.9470) & (0.6644,0.7099) & (0.6687,0.7177) \\
\bottomrule
\end{tabular}
}
\label{tab:cls_binary}
\end{sidewaystable*}

\begin{sidewaystable*}[h]
\centering
\caption{
Detailed results on multiclass classification datasets in medical image diagnosis task.
Results are reported in terms of macro-F1 score.
\textbf{\textcolor{red}{Red}} indicates the overall best results, and \textbf{\textcolor{blue}{Blue}} indicates the overall second-best results.
\textbf{Black} denotes the best results of each model group.
Numbers in parentheses indicate a 95\% confidence interval (CI).
}
\adjustbox{totalheight=0.6\textheight, width=1\linewidth}{
\begin{tabular}{l|p{2.5cm}<{\centering}|p{2.5cm}<{\centering}|p{2.5cm}<{\centering}|p{2.5cm}<{\centering}|p{2.5cm}<{\centering}|p{2.5cm}<{\centering}|p{2.5cm}<{\centering}|p{2.5cm}<{\centering}|p{2cm}<{\centering}}
\toprule
\rowcolor{lightgray}
\textbf{Method} & \textbf{DermNet} & \textbf{OCTMNIST} & \textbf{PathMNIST} & \textbf{Derm7pt} & \textbf{RetOCT} & \textbf{FMC-Endo} & \textbf{OrganAMNIST} & \textbf{HAM10000} & \textbf{Average}\\
\midrule
\rowcolor{lightgray!50}
\multicolumn{10}{c}{\textbf{Generalist Foundation Model} } \\
\multirow{2}{*}{RadFM} & \multirow{2}{*}{N/A} & \multirow{2}{*}{N/A} & \multirow{2}{*}{N/A} & \multirow{2}{*}{N/A} & \multirow{2}{*}{N/A} & \multirow{2}{*}{N/A} & \multirow{2}{*}{N/A} & \multirow{2}{*}{N/A} & \multirow{2}{*}{N/A}\\
 &&&&&&&&&\\
 \hline
\multirow{2}{*}{LLaVA-Med} & 0.0444 & 0.1305 & 0.0834 & 0.1097 & 0.0586 & 0.0813 & 0.0575 & 0.0871 & \multirow{2}{*}{0.0816}\\
 & (0.0413,0.0476) & (0.1219,0.1396) & (0.0803,0.0864) & (0.0975,0.1244) & (0.0552,0.0622) & (0.0764,0.0867) & (0.0520,0.0632) & (0.0806,0.0930)& \\
 \hline
\multirow{2}{*}{Med-Flamingo} & 0.0395 & 0.1000 & \multirow{2}{*}{N/A} & 0.0755 & 0.0278 & 0.0236 & 0.0484 & 0.0140 & \multirow{2}{*}{N/A}\\
 & (0.0366,0.0428) & (0.0953,0.1044) && (0.0703,0.0813) & (0.0265,0.0291) & (0.0199,0.0276) & (0.0476,0.0492) & (0.0127,0.0152) &\\
 \hline
\multirow{2}{*}{InternVL} & 0.1487 & 0.4542 & 0.0783 & 0.1608 & 0.0892 & 0.1169 & 0.0498 & 0.0478 & \multirow{2}{*}{0.1432}\\
 & (0.1437,0.1538) & (0.4427,0.4650) & (0.0758,0.0809) & (0.1344,0.1882) & (0.0856,0.0929) & (0.1112,0.1229) & (0.0483,0.0514) & (0.0413,0.0547) &\\
 \hline
\multirow{2}{*}{MedDr} & \textbf{0.2995} & \textbf{0.5831} & \textbf{0.2306} & \textbf{0.2331} & \textbf{0.7458} & \textbf{0.3197} & \textbf{0.2066} & \textbf{0.3324} & \multirow{2}{*}{\textbf{0.3689}}\\
 & (0.2926,0.3059) & (0.5743,0.5921) & (0.2271,0.2342) & (0.2031,0.2627) & (0.7404,0.7514) & (0.2987,0.3426) & (0.2041,0.2093) & (0.3163,0.3479) &\\
\midrule
\rowcolor{lightgray!50}
\multicolumn{10}{c}{\textbf{Specialist Model} } \\
\multirow{2}{*}{VGG16} & 0.4406 & \textbf{\textcolor{blue}{0.9337}} & 0.9175 & 0.1480 & \textbf{0.8180} & \textbf{\textcolor{blue}{0.3260}} & 0.9377 & 0.4815 & \multirow{2}{*}{0.6254}\\
& (0.4318,0.4489) & (0.9263,0.9413) & (0.9144,0.9206) & (0.1375,0.1634) & (0.8153,0.8209) & (0.3110,0.3473) & (0.9357,0.9396) & (0.4636,0.4981) &\\
\hline
\multirow{2}{*}{AlexNet} & 0.4809 & 0.8594 & 0.9076 & 0.1496 & 0.8020 & 0.2615 & 0.9572 & 0.5688 & \multirow{2}{*}{0.6234}\\
 & (0.4718,0.4896) & (0.8484,0.8690) & (0.9038,0.9113) & (0.1378,0.1626) & (0.7982,0.8056) & (0.2536,0.2703) & (0.9555,0.9589) & (0.5477,0.5900) &\\
 \hline
\multirow{2}{*}{ResNet-18} & 0.4151 & 0.8612 & 0.9186 & 0.1588 & 0.8123 & 0.2801 & 0.9519 & 0.4617 & \multirow{2}{*}{0.6075}\\
 & (0.4070,0.4235) & (0.8495,0.8716) & (0.9147,0.9222) & (0.1486,0.1711) & (0.8090,0.8154) & (0.2728,0.2884) & (0.9501,0.9537) & (0.4417,0.4819) &\\
 \hline
\multirow{2}{*}{DenseNet-121} & 0.5020 & 0.8201 & 0.9453 & 0.1949 & 0.8126 & 0.2855 & 0.9618 & 0.5137 & \multirow{2}{*}{0.6295}\\
 & (0.4931,0.5104) & (0.8084,0.8308) & (0.9421,0.9480) & (0.1791,0.2134) & (0.8095,0.8155) & (0.2754,0.2967) & (0.9601,0.9634) & (0.4943,0.5319) &\\
 \hline
\multirow{2}{*}{EfficientNet} & 0.3638 & 0.8806 & 0.8939 & 0.1267 & 0.7996 & 0.2649 & 0.9532 & 0.3970 & \multirow{2}{*}{0.5850}\\
 & (0.3560,0.3720) & (0.8709,0.8906) & (0.8899,0.8981) & (0.1211,0.1324) & (0.7960,0.8032) & (0.2565,0.2735) & (0.9515,0.9547) & (0.3771,0.4180) &\\
 \hline
\multirow{2}{*}{ViT} & 0.5535 & 0.8707 & 0.9364 & 0.1990 & 0.8037 & 0.2763 & 0.9642 & \textbf{\textcolor{blue}{0.5738}} & \multirow{2}{*}{\textbf{0.6472}}\\
 & (0.5447,0.5618) & (0.8596,0.8807) & (0.9329,0.9397) & (0.1747,0.2254) & (0.7999,0.8072) & (0.2693,0.2830) & (0.9627,0.9655) & (0.5511,0.5959) &\\
 \hline
\multirow{2}{*}{CLIP ViT} & 0.5482 & 0.8678 & 0.9325 & 0.1238 & 0.6385 & 0.1521 & 0.9425 & 0.3799 & \multirow{2}{*}{0.5732}\\
 & (0.5395,0.5566) & (0.8573,0.8776) & (0.9291,0.9358) & (0.1174,0.1299) & (0.6318,0.6454) & (0.1490,0.1566) & (0.9405,0.9442) & (0.3599,0.3990) &\\
 \hline
\multirow{2}{*}{EVA-02 ViT} & 0.4875 & 0.8846 & \textbf{0.9492} & 0.1416 & 0.8046 & 0.1482 & 0.9616 & 0.4969 & \multirow{2}{*}{0.6093}\\
 & (0.4793,0.4958) & (0.8747,0.8947) & (0.9464,0.9522) & (0.1294,0.1576) & (0.8009,0.8081) & (0.1464,0.1500) & (0.9602,0.9631) & (0.4786,0.5173) &\\
 \hline
\multirow{2}{*}{DINO ViT} & 0.5584 & 0.8343 & 0.9225 & \textbf{\textcolor{blue}{0.3004}} & 0.8040 & 0.1829 & 0.9566 & 0.5576 & \multirow{2}{*}{0.6396}\\
 & (0.5494,0.5666) & (0.8227,0.8453) & (0.9193,0.9257) & \textbf{(0.2694,0.3273)} & (0.8005,0.8075) & (0.1772,0.1895) & (0.9551,0.9582) & (0.5351,0.5771) &\\
 \hline
\multirow{2}{*}{SAM ViT} & 0.3451 & 0.9111 & 0.9392 & 0.1323 & 0.7997 & 0.2570 & \textbf{0.9651} & 0.4407 & \multirow{2}{*}{0.5988}\\
 & (0.3374,0.3521) & (0.9022,0.9195) & (0.9358,0.9422) & (0.1262,0.1378) & (0.7957,0.8034) & (0.2478,0.2668) & (0.9635,0.9665) & (0.4226,0.4600) &\\
\midrule
\rowcolor{lightgray!50}
\multicolumn{10}{c}{\textbf{Collaborative Method} } \\
\multirow{2}{*}{Voting} & \textbf{\textcolor{blue}{0.5883}} & 0.8950 & \textbf{\textcolor{red}{0.9704}} & 0.1431 & \textbf{\textcolor{blue}{0.8163}} & 0.2153 & \textbf{\textcolor{red}{0.9792}} & 0.5534 & \multirow{2}{*}{\textbf{\textcolor{blue}{0.6451}}}\\
 & (0.5784,0.5966) & (0.8860,0.9042) & (0.9680,0.9728) & (0.1377,0.1486) & (0.8132,0.8194) & (0.2090,0.2216) & (0.9781,0.9803) & (0.5335,0.5726) &\\
 \hline
\multirow{2}{*}{GSCo} & \textbf{\textcolor{red}{0.5965}} & \textbf{\textcolor{red}{0.9703}} & \textbf{\textcolor{blue}{0.9626}} & \textbf{\textcolor{red}{0.3982}} & \textbf{\textcolor{red}{0.9700}} & \textbf{\textcolor{red}{0.3511}} & \textbf{\textcolor{blue}{0.9749}} & \textbf{\textcolor{red}{0.5768}} & \multirow{2}{*}{\textbf{\textcolor{red}{0.7251}}}\\
 & (0.5881,0.6051) & (0.9648,0.9751) & (0.9600,0.9650) & (0.3572,0.4350) & (0.9668,0.9730) & (0.3273,0.3754) & (0.9737,0.9761) & (0.5551,0.5967) &\\
\bottomrule
\end{tabular}
}
\label{tab:cls_multiclass}
\end{sidewaystable*}

\begin{sidewaystable*}[h]
\centering
\caption{
Detailed results on multilabel classification datasets in medical image diagnosis task.
Results are reported in terms of macro-F1 score.
\textbf{\textcolor{red}{Red}} indicates the overall best results, and \textbf{\textcolor{blue}{Blue}} indicates the overall second-best results.
\textbf{Black} denotes the best results of each group (i.e., Generalist Foundation Model, Specialist Model, or Collaborative Method).
Numbers in parentheses indicate a 95\% confidence interval (CI).
}
\adjustbox{totalheight=0.6\textheight, width=1\linewidth}{
\begin{tabular}{l|p{3.4cm}<{\centering}|p{3.4cm}<{\centering}|p{3.4cm}<{\centering}|p{3.4cm}<{\centering}|p{3.4cm}<{\centering}|p{3.4cm}<{\centering}|p{2cm}<{\centering}}
\toprule
\rowcolor{lightgray}
\textbf{Method} & \textbf{VinDr-PCXR} & \textbf{VinDr-SpineXR} & \textbf{VinDr-Mammo} & \textbf{BRSET} & \textbf{ChestMNIST} & \textbf{FMC-Chest} &\textbf{Average}\\
\midrule
\rowcolor{lightgray!50}
\multicolumn{8}{c}{\textbf{Generalist Foundation Model} } \\
\multirow{2}{*}{RadFM} & 0.0808 & 0.1667 & 0.1389 & \multirow{2}{*}{N/A} & 0.0491 & 0.0148 & \multirow{2}{*}{N/A} \\
 & (0.0782,0.0833) & (0.1636,0.1701) & (0.1373,0.1409) && (0.0476,0.0505) & (0.0137,0.0158) &\\
 \hline
\multirow{2}{*}{LLaVA-Med} & \multirow{2}{*}{N/A} & \multirow{2}{*}{N/A} & \multirow{2}{*}{N/A} & 0.0111 & \multirow{2}{*}{N/A} & 0.0344 & \multirow{2}{*}{N/A} \\
 &&&& (0.0097,0.0128) && (0.0325,0.0363) &\\
 \hline
\multirow{2}{*}{Med-Flamingo} & 0.0231 & 0.0847 & \multirow{2}{*}{N/A} & 0.0081 & \multirow{2}{*}{N/A} & 0.0272 & \multirow{2}{*}{N/A} \\
 & (0.0211,0.0255) & (0.0836,0.0859) && (0.0076,0.0087) && (0.0263,0.0281) &\\
 \hline
\multirow{2}{*}{InternVL} & 0.0695 & 0.1605 & 0.0824 & 0.0214 & 0.0510 & 0.0188 & \multirow{2}{*}{0.0673} \\
 & (0.0667,0.0727) & (0.1546,0.1656) & (0.0805,0.0843) & (0.0183,0.0252) & (0.0503,0.0518) & (0.0169,0.0211) &\\
 \hline
\multirow{2}{*}{MedDr} & \textbf{0.0817} & \textbf{0.2682} & \textbf{0.1935} & \textbf{0.0782} & \textbf{0.1339} & \textbf{0.1082} & \multirow{2}{*}{\textbf{0.1440}} \\
 & (0.0782,0.0856) & (0.2613,0.2752) & (0.1906,0.1965) & (0.0744,0.0827) & (0.1319,0.1361) & (0.1039,0.1124) &\\
\midrule
\rowcolor{lightgray!50}
\multicolumn{8}{c}{\textbf{Specialist Model} } \\
\multirow{2}{*}{VGG16} & 0.0647 & \textbf{\textcolor{blue}{0.3548}} & 0.1766 & \textbf{\textcolor{red}{0.3557}} & 0.1292 & 0.1048 & \multirow{2}{*}{0.1976} \\
 & (0.0613,0.0680) & (0.3433,0.3660) & (0.1752,0.1779) & (0.3469,0.3645) & (0.1268,0.1316) & (0.1026,0.1068) &\\
 \hline
\multirow{2}{*}{AlexNet} & 0.0775 & \textbf{\textcolor{red}{0.3696}} & 0.1729 & 0.3051 & 0.1321 & \textbf{\textcolor{blue}{0.1517}} & \multirow{2}{*}{\textbf{\textcolor{blue}{0.2015}}} \\
 & (0.0729,0.0824) & (0.3573,0.3814) & (0.1714,0.1741) & (0.2922,0.3160) & (0.1295,0.1346) & (0.1472,0.1559) &\\
 \hline
\multirow{2}{*}{ResNet-18} & 0.0644 & 0.2733 & 0.1870 & 0.2462 & 0.1354 & 0.0974 & \multirow{2}{*}{0.1673} \\
 & (0.0613,0.0678) & (0.2634,0.2834) & (0.1856,0.1884) & (0.2391,0.2531) & (0.1329,0.1377) & (0.0949,0.1000) &\\
 \hline
\multirow{2}{*}{DenseNet-121} & \textbf{\textcolor{blue}{0.0795}} & 0.3293 & 0.2037 & 0.3241 & \textbf{\textcolor{blue}{0.1578}} & 0.1022 & \multirow{2}{*}{0.1994} \\
 & (0.0748,0.0843) & (0.3194,0.3390) & (0.1983,0.2092) & (0.3135,0.3331) & (0.1551,0.1607) & (0.0993,0.1050) &\\
 \hline
\multirow{2}{*}{EfficientNet} & 0.0608 & 0.2920 & 0.1957 & 0.2275 & 0.1097 & 0.0766 & \multirow{2}{*}{0.1604} \\
 & (0.0582,0.0638) & (0.2811,0.3037) & (0.1908,0.2010) & (0.2184,0.2371) & (0.1078,0.1116) & (0.0741,0.0793) &\\
 \hline
\multirow{2}{*}{ViT} & 0.0528 & 0.3035 & \textbf{\textcolor{blue}{0.2152}} & 0.3194 & 0.0893 & 0.0778 & \multirow{2}{*}{0.1763} \\
 & (0.0522,0.0534) & (0.2957,0.3115) & (0.2101,0.2206) & (0.3074,0.3311) & (0.0875,0.0910) & (0.0756,0.0799) &\\
 \hline
\multirow{2}{*}{CLIP ViT}& 0.0587 & 0.2902 & 0.1767 & 0.2149 & 0.0850 & 0.0705 & \multirow{2}{*}{0.1493} \\
 & (0.0562,0.0615) & (0.2814,0.2987) & (0.1752,0.1783) & (0.2068,0.2222) & (0.0832,0.0868) & (0.0675,0.0740) &\\
 \hline
\multirow{2}{*}{EVA-02 ViT} & 0.0528 & 0.2448 & 0.0993 & 0.1396 & 0.0932 & 0.0455 & \multirow{2}{*}{0.1125} \\
 & (0.0522,0.0534) & (0.2351,0.2534) & (0.0982,0.1003) & (0.1340,0.1454) & (0.0916,0.0946) & (0.0439,0.0472) &\\
 \hline
\multirow{2}{*}{DINO ViT} & 0.0654 & 0.3158 & 0.1764 & 0.2934 & 0.1201 & 0.0847 & \multirow{2}{*}{0.1760} \\
 & (0.0618,0.0696) & (0.3063,0.3258) & (0.1750,0.1777) & (0.2818,0.3046) & (0.1178,0.1223) & (0.0821,0.0873) &\\
 \hline
\multirow{2}{*}{SAM ViT} & 0.0706 & 0.1892 & 0.1767 & 0.0999 & 0.0673 & 0.0607 & \multirow{2}{*}{0.1107} \\
 & (0.0672,0.0741) & (0.1822,0.1969) & (0.1746,0.1785) & (0.0975,0.1023) & (0.0666,0.0681) & (0.0586,0.0629) &\\
\midrule
\rowcolor{lightgray!50}
\multicolumn{8}{c}{\textbf{Collaborative Method} } \\
\multirow{2}{*}{Voting} & 0.0559 & 0.2853 & 0.1886 & 0.2703 & 0.0927 & 0.0813 & \multirow{2}{*}{0.1624} \\
 & (0.0541,0.0581) & (0.2783,0.2924) & (0.1873,0.1899) & (0.2614,0.2794) & (0.0909,0.0946) & (0.0791,0.0836) &\\
 \hline
\multirow{2}{*}{GSCo} & \textbf{\textcolor{red}{0.0859}} & \textbf{0.2925} & \textbf{\textcolor{red}{0.2311}} & \textbf{\textcolor{blue}{0.3342}} & \textbf{\textcolor{red}{0.2363}} & \textbf{\textcolor{red}{0.1572}} & \multirow{2}{*}{\textbf{\textcolor{red}{0.2229}}} \\
 & (0.0814,0.0905) & (0.2848,0.3014) & (0.2262,0.2364) & (0.3226,0.3451) & (0.2331,0.2396) & (0.1514,0.1634) &\\
\bottomrule
\end{tabular}
}
\label{tab:cls_multilabel}
\end{sidewaystable*}
\begin{figure}[!ht]
\centering
\includegraphics[width=\linewidth, page=1]{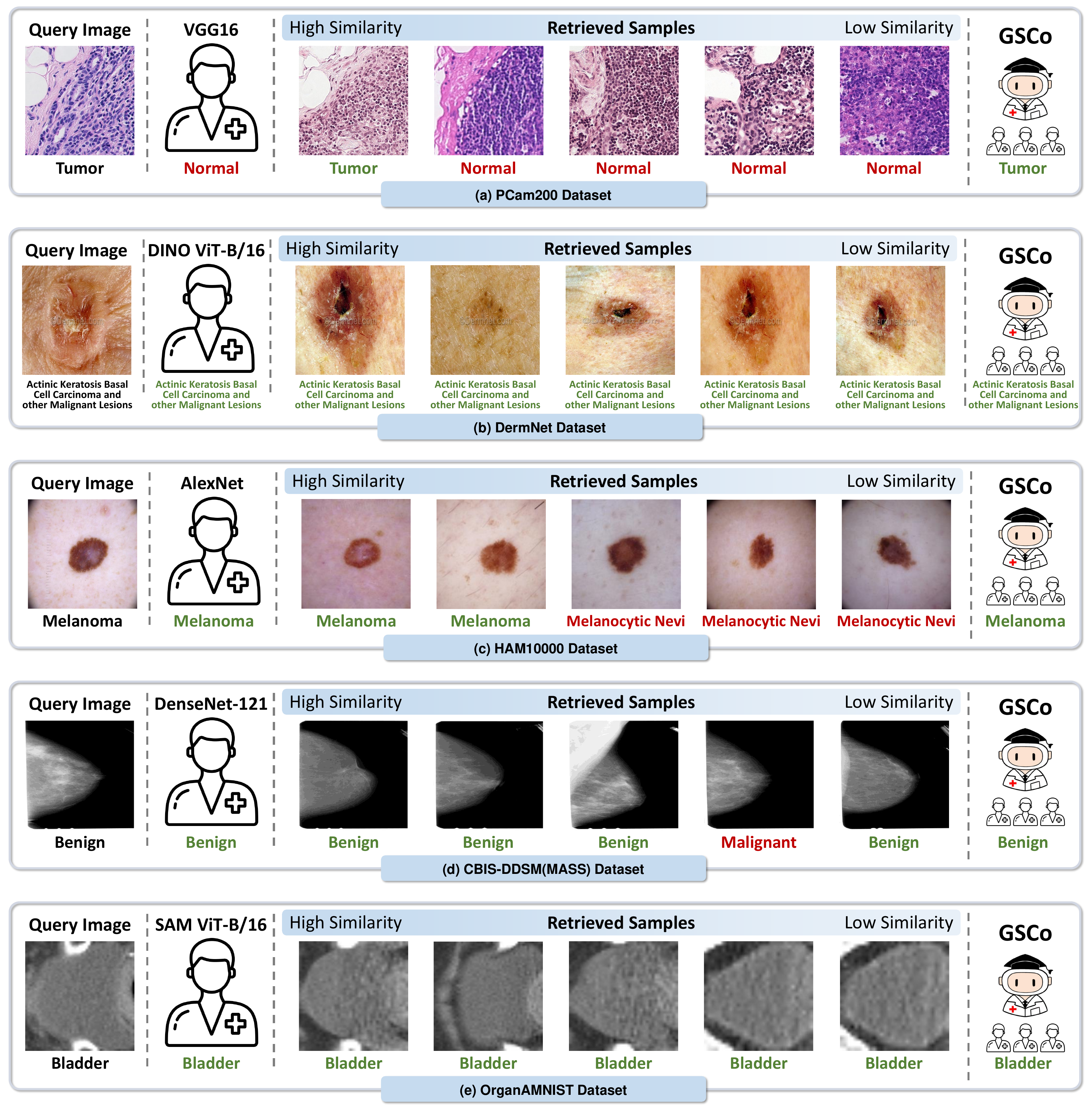}
\caption{
Examples of Retrieval-Augmented Diagnosis on five downstream medical image diagnosis datasets.
From left to right, the contents of each column are: query image and its label, best specialist model and its prediction, Top-$5$ retrieved similar images and their labels, and prediction of GSCo.
\textbf{\textcolor{green1}{Green}} indicates the correct results while \textbf{\textcolor{red1}{Red}} indicates the erroneous results.
}
\label{fig:rad_res_more}
\end{figure}
\begin{figure}[h]
\centering
\includegraphics[width=\linewidth, page=1]{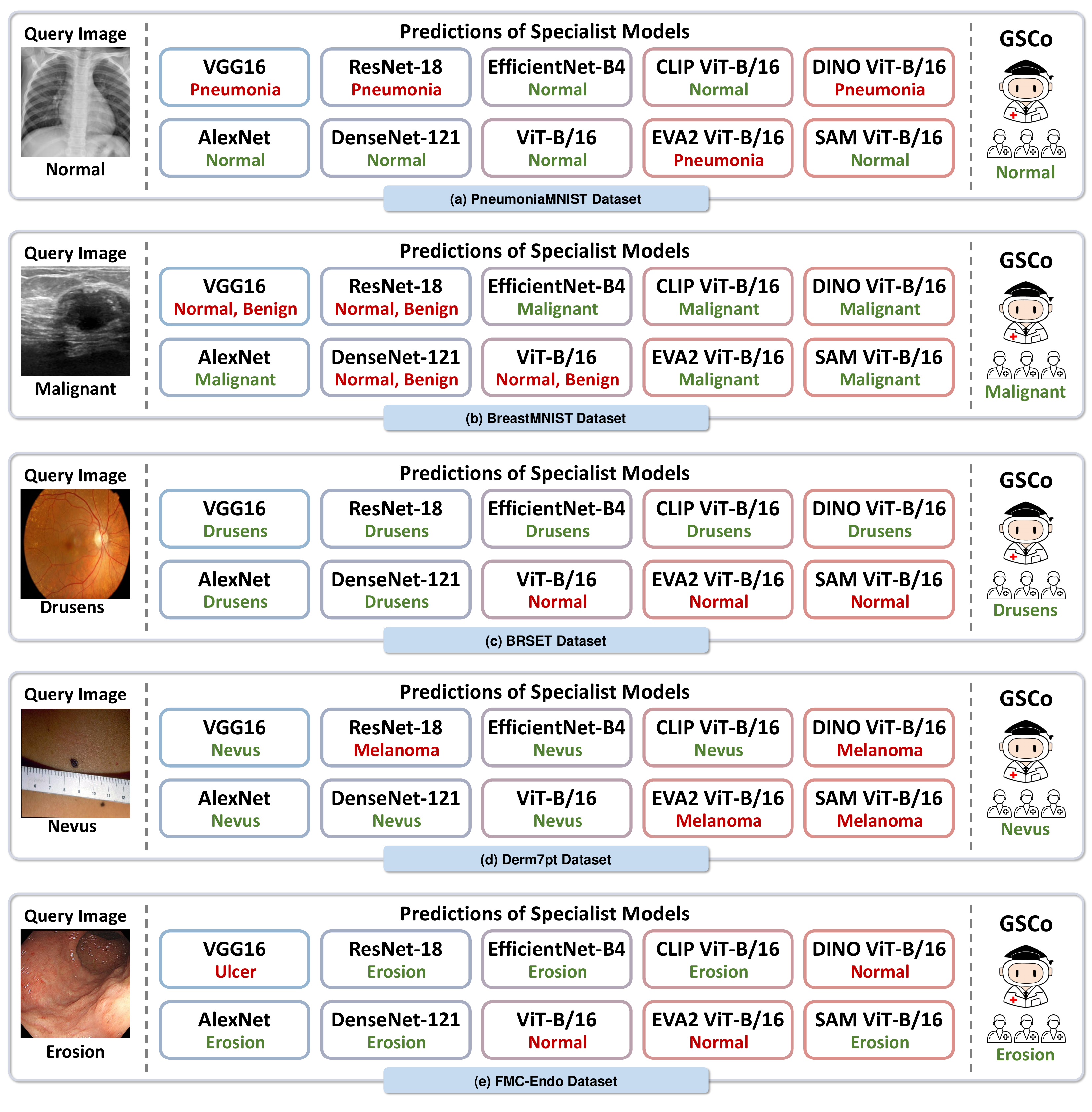}
\caption{
Examples of Mixture-of-Expert Diagnosis on five downstream medical image diagnosis datasets.
Query image and its label, predictions of specialist models, and final diagnosis of GSCo are listed.
\textbf{\textcolor{green1}{Green}} indicates the correct results while \textbf{\textcolor{red1}{Red}} indicates the erroneous results.
}
\label{fig:moed_res_more}
\end{figure}
\begin{table*}
\centering
\caption{
Performance of generalist foundation models on in-domain visual question answering datasets. 
\textbf{\textcolor{red}{Red}} indicates the best results, and \textbf{\textcolor{blue}{Blue}} indicates the second best results.
`*' means that the model is finetuned on the specific dataset.
Numbers in parentheses indicate 95\% CI.
}
\label{tab:indomainvqa}
\adjustbox{width=1\linewidth}{
\begin{tabular}{ll|p{2.2cm}<{\centering}p{2.2cm}<{\centering}p{2.2cm}<{\centering}p{2.2cm}<{\centering}p{2.2cm}<{\centering}}
\toprule
Dataset &Metric &RadFM  &\makecell{LLaVA-\\Med}  &\makecell{Med-\\Flamingo} &InternVL &\ours{}  \\
\midrule
\multirow{12}{*}{VQA-RAD}   
&\multirow{2}{*}{BLEU-1}       &48.64  &37.46*&14.23&\textbf{\textcolor{blue}{55.84}}&\textbf{\textcolor{red}{59.62}}\\
&&(45.21,51.92)&(34.21,40.83)&(12.86,15.52)&(52.36,59.50)&(56.24,63.03)\\
&\multirow{2}{*}{ClosedAcc} &65.34 &22.31*&47.41&\textbf{\textcolor{blue}{74.50}}&\textbf{\textcolor{red}{78.88}}\\ 
&&(61.11,69.86)&(18.71,26.09)&(42.86,52.32)&(70.37,78.57)&(75.30,82.78)\\
&\multirow{2}{*}{OpenRecall}&36.83 &\textbf{\textcolor{red}{63.35}}*&25.73&\textbf{\textcolor{blue}{38.77}}&37.45\\ 
&&(32.40,41.63)&(58.68,68.13)&(21.75,29.81)&(34.16,43.44)&(32.77,42.31)\\
&\multirow{2}{*}{Recall}       &52.70  &38.51*&37.58&\textbf{\textcolor{blue}{58.65}}&\textbf{\textcolor{red}{60.51}}\\ 
&&(49.36,56.30)&(35.26,41.68)&(34.28,40.68)&(55.52,62.11)&(57.14,63.76)\\
&\multirow{2}{*}{OpenAcc}   &\textbf{\textcolor{red}{31.50}} &59.50*&18.00&\textbf{\textcolor{red}{31.50}}&30.00\\
&&(26.67,36.30)&(54.48,64.62)&(13.74,21.95)&(26.67,36.59)&(25.38,34.89)\\
&\multirow{2}{*}{F1}           &50.05 &38.38*&19.98&\textbf{\textcolor{blue}{57.15}}&\textbf{\textcolor{red}{61.10}}\\ 
&&(46.74,53.49)&(34.93,41.58)&(18.25,21.74)&(53.60,60.59)&(58.05,64.34)\\
\hline
\multirow{12}{*}{Slake-VQA}   
&\multirow{2}{*}{BLEU-1}       &\textbf{\textcolor{blue}{75.22}} &66.19*&10.87&45.70&\textbf{\textcolor{red}{76.48}}\\ 
&&(73.35,77.25)&(64.08,68.19)&(10.09,11.66)&(43.42,47.86)&(74.83,78.45)\\
&\multirow{2}{*}{ClosedAcc} &\textbf{\textcolor{blue}{74.08}} &38.87*&46.48&67.61&\textbf{\textcolor{red}{83.38}}\\ 
&&(70.87,77.73)&(35.16,42.86)&(42.79,50.46)&(63.84,71.18)&(80.53,86.28)\\
&\multirow{2}{*}{OpenRecall}&\textbf{\textcolor{blue}{77.26}} &\textbf{\textcolor{red}{81.55}}*&27.19&41.64&74.18\\ 
&&(75.04,79.66)&(79.50,83.58)&(24.82,29.63)&(39.01,44.12)&(71.83,76.47)\\
&\multirow{2}{*}{Recall}       &\textbf{\textcolor{blue}{76.20}} &67.27*&32.80&50.33&\textbf{\textcolor{red}{77.26}}\\ 
&&(74.32,78.04)&(65.17,69.30)&(30.73,34.78)&(48.02,52.57)&(75.32,79.17)\\
&\multirow{2}{*}{OpenAcc}   &\textbf{\textcolor{blue}{74.08}} &\textbf{\textcolor{red}{78.19}}*&22.95&37.68&70.53\\ 
&&(71.70,76.56)&(75.85,80.69)&(20.52,25.17)&(34.92,40.36)&(67.91,73.20)\\
&\multirow{2}{*}{F1}           &\textbf{\textcolor{blue}{75.98}} &66.87*&15.95&47.07&\textbf{\textcolor{red}{77.56}}\\
&&(74.16,77.86)&(64.78,69.06)&(14.81,17.11)&(44.88,49.14)&(75.73,79.47)\\
\hline
\multirow{12}{*}{Path-VQA}   
&\multirow{2}{*}{BLEU-1}       &24.89 &\textbf{\textcolor{blue}{44.79}}*&10.27&32.67&\textbf{\textcolor{red}{61.43}}\\ 
&&(24.07,25.71)&(43.88,45.66)&(10.00,10.55)&(31.87,33.53)&(60.59,62.33)\\
&\multirow{2}{*}{ClosedAcc} &48.86 &56.00*&57.39&\textbf{\textcolor{blue}{62.11}}&\textbf{\textcolor{red}{90.21}}\\ 
&&(47.68,50.12)&(54.74,57.21)&(56.00,58.72)&(60.88,63.35)&(89.48,90.94)\\
&\multirow{2}{*}{OpenRecall}&2.49 &\textbf{\textcolor{red}{37.91}}*&6.64&6.11&\textbf{\textcolor{blue}{33.47}}\\
&&(2.15,2.79)&(36.65,39.10)&(6.10,7.17)&(5.59,6.63)&(32.26,34.61)\\
&\multirow{2}{*}{Recall}       &25.70 &\textbf{\textcolor{blue}{46.07}}*&32.02&34.19&\textbf{\textcolor{red}{61.92}}\\ 
&&(24.84,26.52)&(45.23,46.91)&(31.24,32.86)&(33.38,35.00)&(61.09,62.78)\\
&\multirow{2}{*}{OpenAcc}   &1.28 &\textbf{\textcolor{red}{34.87}}*&4.57&3.80&\textbf{\textcolor{blue}{30.50}}\\ 
&&(0.99,1.54)&(33.56,36.13)&(4.00,5.14)&(3.31,4.29)&(29.27,31.68)\\
&\multirow{2}{*}{F1}           &25.24 &\textbf{\textcolor{blue}{45.47}}*&15.22&33.20&\textbf{\textcolor{red}{62.15}}\\
&&(24.50,26.03)&(44.57,46.35)&(14.78,15.60)&(32.35,34.05)&(61.35,63.03)\\
\hline
\multirow{12}{*}{PMC-VQA}   
&\multirow{2}{*}{BLEU-1}       &\textbf{\textcolor{red}{17.74}} &6.65&9.54&8.68&\textbf{\textcolor{blue}{15.22}}\\
&&(17.59,17.90)&(6.55,6.74)&(9.46,9.62)&(8.6,8.75)&(15.09,15.34)\\
&\multirow{2}{*}{ClosedAcc} &\textbf{\textcolor{red}{79.55}} &51.54&26.89&40.06&\textbf{\textcolor{blue}{61.30}}\\ 
&&(76.34,82.83)&(47.64,55.19)&(23.18,30.42)&(36.16,43.72)&(57.27,65.20)\\
&\multirow{2}{*}{OpenRecall}&20.05 &8.09&16.30&\textbf{\textcolor{blue}{23.67}}&\textbf{\textcolor{red}{27.17}}\\ 
&&(19.90,20.21)&(7.98,8.21)&(16.15,16.44)&(23.51,23.84)&(26.99,27.35)\\
&\multirow{2}{*}{Recall}       &20.28 &8.26&16.34&\textbf{\textcolor{blue}{23.73}}&\textbf{\textcolor{red}{27.30}}\\ 
&&(20.11,20.43)&(8.15,8.37)&(16.19,16.48)&(23.57,23.89)&(27.12,27.47)\\
&\multirow{2}{*}{OpenAcc}   &11.05 &3.83&8.08&\textbf{\textcolor{blue}{11.34}}&\textbf{\textcolor{red}{14.94}}\\ 
&&(10.90,11.20)&(3.74,3.93)&(7.94,8.21)&(11.19,11.50)&(14.75,15.13)\\
&\multirow{2}{*}{F1}           &\textbf{\textcolor{red}{20.02}} &7.98&12.03&12.09&\textbf{\textcolor{blue}{18.90}}\\
&&(19.86,20.19)&(7.88,8.09)&(11.93,12.14)&(12.00,12.17)&(18.76,19.02)\\
\bottomrule
\end{tabular}}
\end{table*}

\begin{table*}
\centering
\caption{
Performance of generalist foundation models on the VQA-Med dataset. 
\textbf{\textcolor{red}{Red}} indicates the best results, and \textbf{\textcolor{blue}{Blue}} indicates the second best results.
Numbers in parentheses indicate 95\% CI.
}
\label{tab:outofdomainvqa}
\adjustbox{width=1\linewidth}{
\begin{tabular}{ll|p{2.2cm}<{\centering}p{2.2cm}<{\centering}p{2.2cm}<{\centering}p{2.2cm}<{\centering}p{2.2cm}<{\centering}}
\toprule
Dataset &Metric &RadFM  &\makecell{LLaVA-\\Med}  &\makecell{Med-\\Flamingo} &InternVL &\ours{}  \\
\midrule
\multirow{12}{*}{VQA-Med}   
&\multirow{2}{*}{BLEU-1}       &16.40 &20.91&7.65&\textbf{\textcolor{blue}{25.61}}&\textbf{\textcolor{red}{28.87}}\\
&&(14.87,17.99)&(14.28,19.27)&(6.81,8.41)&(22.89,28.17)&(26.07,31.96)\\
&\multirow{2}{*}{ClosedAcc} &\textbf{\textcolor{blue}{57.92}} &34.38&51.39&57.75&\textbf{\textcolor{red}{73.42}}\\ 
&&(48.78,66.68)&(25.63,43.24)&(41.17,60.54)&(48.57,67.50)&(64.86,81.58)\\
&\multirow{2}{*}{OpenRecall}&\textbf{\textcolor{red}{27.49}} &20.98&18.75&22.57&\textbf{\textcolor{blue}{23.56}}\\ 
&&(24.27,30.24)&(18.19,23.76)&(16.31,2123)&(19.73,25.40)&(20.77,26.19)\\
&\multirow{2}{*}{Recall}       &\textbf{\textcolor{red}{31.41}} &22.56&23.02&27.16&\textbf{\textcolor{blue}{29.96}}\\ 
&&(28.34,34.35)&(16.06,20.92)&(20.47,25.62)&(24.30,29.97)&(27.03,32.85)\\
&\multirow{2}{*}{OpenAcc}   &\textbf{\textcolor{red}{25.11}} &18.13&15.69&20.22&\textbf{\textcolor{blue}{20.96}}\\ 
&&(21.93,28.03)&(15.44,20.64)&(13.09,18.18)&(17.49,23.00)&(18.08,23.68)\\
&\multirow{2}{*}{F1}           &\textbf{\textcolor{blue}{21.70}} &22.35&11.16&26.82&\textbf{\textcolor{red}{30.51}}\\
&&(19.70,23.86)&(16.01,20.78)&(9.91,12.33)&(23.98,29.68)&(27.59,33.35)\\
\bottomrule
\end{tabular}}
\end{table*}

\begin{table*}[ht]
\centering
\caption{
Performance of generalist foundation models on the OmniMedVQA dataset.
Results are reported in terms of accuracy.
\textbf{\textcolor{red}{Red}} indicates the best results, and \textbf{\textcolor{blue}{Blue}} indicates the second best results.
}
\adjustbox{width=1\linewidth}{
\begin{tabular}{lp{1.1cm}<{\centering}p{1.1cm}<{\centering}p{1.1cm}<{\centering}p{1.1cm}<{\centering}p{1.6cm}<{\centering}p{1.6cm}<{\centering}p{1.1cm}<{\centering}p{1.1cm}<{\centering}}
\toprule
Method & MRI & CT & X-Ray & OCT & Microscopy & Dermoscopy & Fundus & Overall \\ \midrule
RadFM & 34.37 & 34.55 & 38.07 & 34.46 & 40.16 & 30.62 & 10.76 & 34.48
\\
LLaVA-Med & 25.35 & 27.79 & 33.97 & 37.71 & 25.34 & 23.64 & 30.92&27.05
 \\
Med-Flamingo & 25.56 & 28.83 & 22.11 & 23.53 & 30.11 & 25.31 & 17.48& 26.19
\\
InternVL & \textbf{\textcolor{blue}{63.23}} & \textbf{\textcolor{blue}{39.59}} & \textbf{\textcolor{blue}{59.79}} & \textbf{\textcolor{blue}{46.57}} & \textbf{\textcolor{blue}{57.22}} & \textbf{\textcolor{blue}{51.87}} & \textbf{\textcolor{blue}{67.06}}& \textbf{\textcolor{blue}{55.61}}
\\
MedDr & \textbf{\textcolor{red}{63.58}} & \textbf{\textcolor{red}{61.50}} & \textbf{\textcolor{red}{71.98}} & \textbf{\textcolor{red}{56.52}} & \textbf{\textcolor{red}{59.71}} & \textbf{\textcolor{red}{57.08}} & \textbf{\textcolor{red}{80.67}}& \textbf{\textcolor{red}{62.98}}
\\  
\bottomrule
\end{tabular}}
\label{tab:omnimed_vqa}
\end{table*}

\begin{table*}[ht]
\centering
\caption{
Ablation Study of the GSCo framework.
For binary classification, accuracy is reported. 
Otherwise, the Macro-F1 score is reported.
\textbf{\textcolor{red}{Red}} indicates the best results, and \textbf{\textcolor{blue}{Blue}} indicates the second best results.
Numbers in parentheses indicate the improvement relative to \ours{}. 
}
\adjustbox{width=1\linewidth}{
\begin{tabular}{l|p{1cm}<{\centering}p{2.7cm}<{\centering}p{2.7cm}<{\centering}p{2.7cm}<{\centering}}
\toprule
Dataset & MedDr & MedDr+MoED & MedDr+RAD & GSCo \\ 
\midrule
PCam200 & 0.9089 & \textbf{\textcolor{blue}{0.9305}} (0.0216$\uparrow$) & 0.9240 (0.0151$\uparrow$) & \textbf{\textcolor{red}{0.9321}} (0.0232$\uparrow$) \\
DermNet & 0.2995 & \textbf{\textcolor{blue}{0.5788}} (0.2793$\uparrow$) & 0.5543 (0.2548$\uparrow$) & \textbf{\textcolor{red}{0.5965}} (0.2970$\uparrow$) \\
HAM10000 & 0.3324 & 0.5534 (0.2210$\uparrow$) & \textbf{\textcolor{blue}{0.5754}} (0.2430$\uparrow$) & \textbf{\textcolor{red}{0.5768}} (0.2444$\uparrow$) \\
RetOCT & 0.7458 & 0.9508 (0.2050$\uparrow$) & \textbf{\textcolor{blue}{0.9671}} (0.2213$\uparrow$) & \textbf{\textcolor{red}{0.9700}} (0.2242$\uparrow$) \\
VinDr-SpineXR & 0.2682 & 0.2775 (0.0093$\uparrow$) & \textbf{\textcolor{blue}{0.2915}} (0.0233$\uparrow$) & \textbf{\textcolor{red}{0.2925}} (0.0243$\uparrow$) \\
PneumoniaMNIST & 0.8734 & \textbf{\textcolor{blue}{0.9535}} (0.0801$\uparrow$) & 0.9327 (0.0593$\uparrow$) & \textbf{\textcolor{red}{0.9599}} (0.0865$\uparrow$) \\
BreastMNIST & 0.7179 & \textbf{\textcolor{blue}{0.8974}} (0.1795$\uparrow$) & \textbf{\textcolor{blue}{0.8974}} (0.1795$\uparrow$) & \textbf{\textcolor{red}{0.9295}} (0.2116$\uparrow$) \\
CBIS-DDSM(CALC) & 0.5184 & 0.6503 (0.1319$\uparrow$) & \textbf{\textcolor{blue}{0.6718}} (0.1534$\uparrow$) & \textbf{\textcolor{red}{0.6933}} (0.1749$\uparrow$) \\
PathMNIST & 0.2306 & 0.9301 (0.6995$\uparrow$) & \textbf{\textcolor{blue}{0.9394}} (0.7088$\uparrow$) & \textbf{\textcolor{red}{0.9626}} (0.7320$\uparrow$) \\
\hline
Average & 0.5439 & 0.7469 (0.2030$\uparrow$) & \textbf{\textcolor{blue}{0.7504}} (0.2065$\uparrow$) & \textbf{\textcolor{red}{0.7681}} (0.2242$\uparrow$) \\
\bottomrule
\end{tabular}
}
\label{tab:cls_ablation}
\end{table*}
\begin{table*}[t]
\centering
\caption{
Performance of generalist foundation models on the medical report generation task.
\textbf{\textcolor{red}{Red}} indicates the best results, and \textbf{\textcolor{blue}{Blue}} indicates the second best results.
Numbers in parentheses indicate 95\% CI.
}
\adjustbox{width=1\linewidth}{
\begin{tabular}{ll|p{2.2cm}<{\centering}p{2.2cm}<{\centering}p{2.2cm}<{\centering}p{2.2cm}<{\centering}p{2.2cm}<{\centering}}
\toprule
Dataset &Metric &RadFM  &\makecell{LLaVA-\\Med}  &\makecell{Med-\\Flamingo} &InternVL &\ours{}  \\
\midrule
\multirow{14}{*}{MIMIC-CXR} 
&\multirow{2}{*}{F1-RadGraph}   &\textbf{\textcolor{blue}{18.25}}  &6.36&7.13&6.70&\textbf{\textcolor{red}{22.44}}\\ 
&&(17.95,18.55)&(6.22,6.50)&(6.96,7.29)&(6.56,6.86)&(22.13,22.74)\\
&\multirow{2}{*}{BLEU-1}        &22.19  &19.34&22.43&\textbf{\textcolor{blue}{25.58}}&\textbf{\textcolor{red}{27.28}}\\
&&(21.83,22.57)&(19.09,19.60)&(22.24,22.64)&(25.30,25.87)&(26.83,27.73)\\
&\multirow{2}{*}{BLEU-4}        &\textbf{\textcolor{blue}{5.55}}  &0.96&1.91&1.69&\textbf{\textcolor{red}{7.59}}\\ 
&&(5.39,5.74)&(0.92,1.01)&(1.85,1.98)&(1.63,1.75)&(7.39,7.78)\\
&\multirow{2}{*}{ROUGE-1}       &\textbf{\textcolor{blue}{28.88}}  &21.44&21.69&22.87&\textbf{\textcolor{red}{32.58}}\\ 
&&(28.62,29.09)&(21.30,21.57)&(21.54,21.85)&(22.70,23.04)&(32.31,32.84)\\
&\multirow{2}{*}{ROUGE-L}       &\textbf{\textcolor{blue}{20.52}}  &13.90&14.60&15.61&\textbf{\textcolor{red}{22.59}}\\ 
&&(20.31,20.75)&(13.82,13.98)&(14.51,14.68)&(15.51,15.71)&(22.38,22.79)\\
&\multirow{2}{*}{CheXbert Vec} &\textbf{\textcolor{blue}{31.18}} &15.56&18.69&15.90&\textbf{\textcolor{red}{34.18}}\\ 
&&(30.63,31.73)&(15.19,15.90)&(18.30,19.06)&(15.56,16.26)&(33.69,34.65)\\
&\multirow{2}{*}{METEOR}  &\textbf{\textcolor{blue}{20.42}}&13.25&14.02&17.54&\textbf{\textcolor{red}{23.77}}\\
&&(20.19,20.64)&(13.16,13.34)&(13.93,14.10)&(17.44,17.65)&(23.54,24.05)\\
\midrule
\multirow{14}{*}{IU-Xray} 
&\multirow{2}{*}{F1-RadGraph}       &\textbf{\textcolor{blue}{29.17}}&4.72&11.91&11.21&\textbf{\textcolor{red}{33.19}}\\ 
&&(28.62,29.69)&(4.45,4.97)&(11.45,12.36)&(10.81,11.60)&(32.64,33.75)\\
&\multirow{2}{*}{BLEU-1}            &\textbf{\textcolor{red}{40.78}}&14.85&16.28&18.27&\textbf{\textcolor{blue}{37.71}}\\ 
&&(40.32,41.30)&(14.56,15.14)&(15.94,16.60)&(17.93,18.63)&(36.97,38.50)\\
&\multirow{2}{*}{BLEU-4}            &\textbf{\textcolor{blue}{10.28}}&0.68&2.06&1.73&\textbf{\textcolor{red}{12.22}}\\ 
&&(9.95,10.61)&(0.57,0.77)&(1.93,2.18)&(1.63,1.83)&(11.83,12.65)\\
&\multirow{2}{*}{ROUGE-1}           &\textbf{\textcolor{blue}{36.79}}&12.98&14.93&18.22&\textbf{\textcolor{red}{39.49}}\\ 
&&(36.32,37.23)&(12.69,13.29)&(14.60,15.26)&(17.90,18.56)&(38.99,39.94)\\
&\multirow{2}{*}{ROUGE-L}           &\textbf{\textcolor{blue}{26.07}}&9.96&11.11&13.43&\textbf{\textcolor{red}{28.35}}\\ 
&&(25.72,26.42)&(9.74,10.19)&(10.88,11.34)&(13.20,13.66)&(27.91,28.79)\\
&\multirow{2}{*}{CheXbert Vec}   &\textbf{\textcolor{red}{59.26}}&14.01&24.25&24.17&\textbf{\textcolor{blue}{56.47}}\\ 
&&(58.26,60.21)&(13.47,14.57)&(23.47,24.98)&(23.48,24.82)&(55.71,57.34)\\
&\multirow{2}{*}{METEOR}            &\textbf{\textcolor{blue}{30.92}}&14.48&17.69&20.03&\textbf{\textcolor{red}{32.34}}\\ 
&&(30.44,31.37)&(14.27,14.72)&(17.37,18.02)&(19.74,20.29)&(31.82,32.92)\\
\bottomrule
\end{tabular}}
\label{tab:mrg} 
\end{table*}
                                                                    
\begin{table*}
\centering
\caption{Performance with retrieval-augmented diagnosis.
\textbf{\textcolor{red}{Red}} indicates the best results, and \textbf{\textcolor{blue}{Blue}} indicates the second best results.
Numbers in parentheses indicate 95\% CI.
}
\adjustbox{width=1\linewidth}{
\begin{tabular}{ll|p{2.2cm}<{\centering}|p{2.2cm}<{\centering}p{2.2cm}<{\centering}|p{2.2cm}<{\centering}p{2.2cm}<{\centering}}
\toprule
\multirow{2}{*}{Dataset} &\multirow{2}{*}{Metric}&\multirow{2}{*}{Voting}&\multicolumn{2}{c|}{Med-Flamingo}  & \multicolumn{2}{c}{\ours{}}   \\
&&&\multicolumn{1}{c}{w/o RAD} & \multicolumn{1}{c|}{w/ RAD} & \multicolumn{1}{c}{w/o RAD} & \multicolumn{1}{c}{w/ RAD} \\
\midrule
&\multirow{2}{*}{Accuracy}     &\textbf{\textcolor{blue}{91.03}}&71.31&85.26&87.34&\textbf{\textcolor{red}{92.95}}\\
Pneumonia-&&(89.87,92.13)&(69.57,73.09)&(83.89,86.59)&(86.06,88.62)&(91.91,93.88)\\
MNIST&\multirow{2}{*}{Macro-F1}     &\textbf{\textcolor{blue}{93.15}}&76.60&89.33&89.16&\textbf{\textcolor{red}{94.18}}\\
&&(92.26,94.02)&(74.89,78.19)&(88.16,90.35)&(88.01,90.28)&(93.29,94.97)\\
\midrule
&\multirow{2}{*}{Accuracy}     &\textbf{\textcolor{blue}{83.97}}&30.77&58.33&71.79&\textbf{\textcolor{red}{87.82}}\\
Breast-&&(81.01,86.56)&(27.22,34.38)&(54.52,62.22)&(68.35,75.41)&(85.16,90.30)\\
MNIST&\multirow{2}{*}{Macro-F1}     &63.77&10.00&55.78&\textbf{\textcolor{red}{80.00}}&\textbf{\textcolor{blue}{76.54}}\\
&&(56.91,70.27)&(6.75,13.99)&(51.05,60.33)&(77.20,82.82)&(71.53,81.08)\\
\midrule
&\multirow{2}{*}{Accuracy}     &\textbf{\textcolor{blue}{67.60}}&25.00&49.40&67.50&\textbf{\textcolor{red}{69.20}}\\
OCT-&&(66.55,68.64)&(25.00,25.00)&(48.81,50.00)&(66.71,68.27)&(68.10,70.37)\\
MNIST&\multirow{2}{*}{Macro-F1}     &62.77&10.00&36.34&\textbf{\textcolor{blue}{58.31}}&\textbf{\textcolor{red}{66.06}}\\
&&(61.35,64.23)&(9.53,10.44)&(35.25,37.40)&(57.43,59.21)&(64.66,67.48)\\
\midrule
\multirow{12}{*}{Slake-VQA}   
&\multirow{2}{*}{BLEU-1}       &- &0.108&21.30&\textbf{\textcolor{blue}{76.48}}&\textbf{\textcolor{red}{77.36}}\\
&&&(10.09,11.66)&(20.31,22.35)&(74.83,78.45)&(75.58,79.19)\\
&\multirow{2}{*}{ClosedAcc} &- &46.48&63.66&\textbf{\textcolor{blue}{83.38}}&\textbf{\textcolor{red}{86.20}}\\ 
&&&(42.79,50.46)&(60.00,67.66)&(80.53,86.28)&(83.63,88.94)\\
&\multirow{2}{*}{OpenRecall}&- &27.19&36.50&\textbf{\textcolor{blue}{74.18}}&\textbf{\textcolor{red}{74.31}}\\ 
&&&(24.82,29.63)&(33.96,39.10)&(71.83,76.47)&(72.00,76.60)\\
&\multirow{2}{*}{Recall}       &- &32.80&45.59&\textbf{\textcolor{blue}{77.26}}&\textbf{\textcolor{red}{78.29}}\\
&&&(30.73,34.78)&(43.27,47.78)&(75.32,79.17)&(76.42,80.08)\\
&\multirow{2}{*}{OpenAcc}   &- &22.95&31.30&\textbf{\textcolor{red}{70.53}}&\textbf{\textcolor{blue}{70.25}}\\ 
&&&(20.52,25.17)&(28.70,33.85)&(67.91,73.20)&(67.83,73.11)\\
&\multirow{2}{*}{F1}         &- &15.95&28.56&\textbf{\textcolor{blue}{77.56}}&\textbf{\textcolor{red}{78.54}}\\
&&&(14.81,17.11)&(27.18,29.90)&(75.73,79.47)&(76.77,80.36)\\
\midrule
\multirow{16}{*}{IU-Xray} 
&\multirow{2}{*}{F1-RadGraph}      
&31.08&11.91&30.35&\textbf{\textcolor{blue}{33.19}}&\textbf{\textcolor{red}{35.21}}\\ 
&&(30.27,31.85)&(11.45,12.36)&(29.58,31.06)&(32.64,33.75)&(34.54,35.88)\\
&\multirow{2}{*}{BLEU-1}           &40.68&16.28&37.65&\textbf{\textcolor{blue}{37.71}}&\textbf{\textcolor{red}{41.80}}\\ 
&&(39.93,41.49)&(15.94,16.60)&(36.86,38.48)&(36.97,38.50)&(40.94,42.63)\\
&\multirow{2}{*}{BLEU-4}           &\textbf{\textcolor{blue}{12.40}}&2.06&11.34&12.22&\textbf{\textcolor{red}{13.75}}\\ 
&&(11.63,13.15)&(1.93,2.18)&(10.63,12.01)&(11.83,12.65)&(13.11,14.36)\\
&\multirow{2}{*}{ROUGE-1}          &36.74&14.93&35.44&\textbf{\textcolor{blue}{39.49}}&\textbf{\textcolor{red}{40.54}}\\
&&(36.03,37.36)&(14.60,15.26)&(34.74,36.07)&(38.99,39.94)&(39.88,41.10)\\
&\multirow{2}{*}{ROUGE-L}          &26.82&11.11&25.83&\textbf{\textcolor{blue}{28.35}}&\textbf{\textcolor{red}{29.54}}\\
&&(26.14,27.49)&(10.88,11.34)&(25.18,26.45)&(27.91,28.79)&(28.97,30.09)\\
&\multirow{2}{*}{CheXbert Vec}  &53.57&24.25&53.16&\textbf{\textcolor{blue}{56.47}}&\textbf{\textcolor{red}{57.50}}\\
&&(52.65,54.44)&(23.47,24.98)&(52.22,54.05)&(55.71,57.34)&(56.60,58.44)\\
&\multirow{2}{*}{METEOR}           &\textbf{\textcolor{blue}{34.05}}&17.69&33.43&32.34&\textbf{\textcolor{red}{34.09}}\\
&&(33.28,34.82)&(17.37,18.02)&(32.71,34.17)&(31.82,32.92)&(33.41,34.78)\\
\bottomrule
\end{tabular}}
\label{tab:rag} 
\end{table*}

\begin{table*}
\centering
\caption{Detailed information of the in-domain medical image diagnosis datasets. The dataset name, classification type, dataset size, and label set are listed.}
\adjustbox{width=1\linewidth}{
\begin{tabular}{l|c|c|p{8cm}}
\toprule
\textbf{Dataset} & \textbf{Type} & \textbf{Size} & \textbf{Label Set} \\
\midrule
PCam200~\cite{kawai2023large}&Binary&17,932&Normal, Tumor.\\
\hline
DermNet~\cite{kaggleDermnet}&Multiclass&4,045&Actinic Keratosis Basal Cell Carcinoma and other Malignant Lesions, Atopic Dermatitis, Warts Molluscum and other Viral Infections, Vascular Tumors, Seborrheic Keratoses and other Benign Tumors, Urticaria Hives, Light Diseases and Disorders of Pigmentation, Exanthems and Drug Eruptions, Scabies Lyme Disease and other Infestations and Bites, Psoriasis pictures Lichen Planus and related diseases, Eczema, Poison Ivy  and other Contact Dermatitis, Acne and Rosacea, Bullous Disease, Hair Loss  Alopecia and other Hair Diseases, Tinea Ringworm Candidiasis and other Fungal Infections, Vasculitis, Lupus and other Connective Tissue diseases, Cellulitis Impetigo and other Bacterial Infections, Melanoma Skin Cancer Nevi and Moles, Systemic Disease, Herpes HPV and other STDs, Nail Fungus and other Nail Disease.\\
\hline
HAM10000~\cite{tschandl2018ham10000}&Multiclass&1,511&Actinic keratoses and intraepithelial carcinoma, Basal cell carcinoma, Benign keratosis-like lesions, Dermatofibroma, Melanoma, Melanocytic nevi, Vascular lesions.\\
\hline
RetOCT~\cite{9740985}&Multiclass&2,800&Age-related macular degeneration, Choroidal neovascularization, Central serous retinopathy, Diabetic macular edema, Macular hole, Drusen, Diabetic retinopathy, Normal.\\
\hline
ChestMNIST~\cite{wang2017chestx}&Multilabel&22,433&Atelectasis, Cardiomegaly, Effusion, Infiltration, Mass, Nodule, Pneumonia, Pneumothorax, Consolidation, Edema, Emphysema, Fibrosis, Pleural, Hernia, No finding\\
\hline
VinDr-SpineXR~\cite{nguyen2021vindr}&Multilabel&2,077&Osteophytes, Vertebral collapse, Disc space narrowing, Surgical implant, Other lesions, Foraminal stenosis, Spondylolysthesis, No finding.\\
\hline
VinDr-PCXR~\cite{pham2022vindr}&Multilabel&1,397&Pneumonia, Bronchiolitis, Other disease, Bronchitis, Brocho-pneumonia, Tuberculosis, Pleuro-pneumonia, Situs inversus, Mediastinal tumor, Diagphramatic hernia, Hyaline membrane disease, Lung tumor, Congenital emphysema, CPAM, No finding.\\
\hline
VinDr-Mammo~\cite{nguyen2023vindr}&Multilabel&4,000&Birads negative, Breast heterogeneously density, Breast scattered areas of fibroglandular, Birads suspicious malignant, Mass, Breast extremely density, Birads benign, Birads highly suggestive of malignant, Suspicious Calcification, Breast almost entirely fatty, Suspicious Lymph Node, Focal Asymmetry, Birads probably benign, Asymmetry, Architectural Distortion, Skin Thickening, Global Asymmetry, Nipple Retraction, Skin Retraction, No Finding.\\
\bottomrule
\end{tabular}
}
\label{tab:indomain_data_stat} 
\end{table*}

\begin{table*}
\centering
\caption{Detailed information of the out-of-domain medical image diagnosis datasets. The dataset name, classification type, dataset size, and label set are listed.}
\adjustbox{width=1\linewidth}{
\begin{tabular}{l|c|c|p{6cm}}
\toprule
\textbf{Dataset} & \textbf{Type} & \textbf{Size} & \textbf{Label Set} \\
\midrule
PneumoniaMNIST~\cite{yang2023medmnist}&Binary&624&Normal, Pneumonia.\\
\hline
BreastMNIST~\cite{yang2023medmnist}&Binary&156&Normal and Benign, Malignant.\\
\hline
OrganAMNIST~\cite{yang2023medmnist}&Multiclass&17,778&Bladder, Femur-left, Femur-right, Heart, Kidney-left, Kidney-right, Liver, Lung-left, Lung-right, Pancreas, Spleen.\\
\hline
PathMNIST~\cite{yang2023medmnist}&Multiclass&7,180&Adipose, Background, Debris, Lymphocytes, Mucus, Colorectal adenocarcinoma epithelium, Smooth muscle, Normal colon mucosa, Cancer-associated stroma.\\
\hline
OCTMNIST~\cite{yang2023medmnist}&Multiclass&1,000&Choroidal neovascularization, Diabetic macular edema, Drusen, Normal.\\
\hline
CBIS-DDSM(CALC)~\cite{sawyer2016curated}&Multiclass&326&Benign, Malignant.\\
\hline
CBIS-DDSM(MASS)~\cite{sawyer2016curated}&Multiclass&378&Benign, Malignant.\\
\hline
FMC-Colon~\cite{wang2023real}&Binary&4,355&Negative, Positive.\\
\hline
FMC-Endo~\cite{wang2023real}&Multiclass&2,055&Ulcer, Erosion, Polyp, Tumor, Normal.\\
\hline
FMC-Chest~\cite{wang2023real}&Multilabel&2,708&Effusion, Nodule, Pneumonia, Cardiomegaly, Hilar enlargement, Fracture old, Fibrosis, Aortic calcification, Tortuous aorta, Thickened pleura, Tb, Pneumothorax, Emphysema, Atelectasis, Calcification, Pulmonary edema, Increased lung markings, Elevated diaphragm, Consolidation, No finding.\\
\hline
Derm7pt~\cite{kawahara2018seven}&Multiclass&395&Nevus, Miscellaneous, Melanoma, Vascular lesion, Lentigo, Melanosis, Dermatofibroma, Basal cell carcinoma, Seborrheic keratosis.\\
\hline
BRSET~\cite{nakayama2023brazilian}&Multilabel&3,254&Diabetic, Macular edema, Scar, Nevus, Age-related macular degeneration, Vascular occlusion, Hypertensive retinopathy, Drusens, Hemorrhage, Retina detachment, Myopia, Increased cup disc, Disease, Normal.\\
\bottomrule
\end{tabular}
}
\label{tab:outdomain_data_stat} 
\end{table*}

\begin{figure}[!ht]
\begin{promptenv}{Instruction for visual question answering.}
You are a helpful medical assistant. \\
You are required to answer the question based on the medical image. \\
The question is \textbf{\{Question\}}.
\end{promptenv}
\begin{promptenv}{Instruction for medical report generation.}
You are a helpful medical assistant. \\
Your task is report generation. \\
You are given a chest x-ray image, and you are required to generate a summary report about the image.
\end{promptenv}
\begin{promptenv}{Instruction for medical image diagnosis.}
You are a helpful medical assistant. \\
Your task is disease diagnosis. \\
You are given a \textbf{\{Modality\}} image.\\
The possible diagnoses are: \textbf{\{Label Set\}}.
\end{promptenv}
\begin{promptenv}{Instruction for diagnosis-guided bootstrapping.}
You are a helpful medical assistant. \\
Your task is report generation. \\
You are given a \textbf{\{Modality\}} image.\\
You need to provide a medical report consisting of findings and impressions.\\
Findings list the observations and impressions outlines the final diagnosis.
\end{promptenv}
\begin{promptenv}{Instruction for medical image description.}
You are a helpful medical assistant. \\
Your task is report generation. \\
You are given a \textbf{\{Modality\}} image.\\
You are required to generate a detailed description and analysis from a medical perspective based on the image.
\end{promptenv}
\caption{
Prompt templates for different instruction-tuning datasets.
}
\label{fig:sftprompt}
\end{figure}
\begin{figure}[!ht]
\begin{promptenv}{Instructions for Generalist-Specialist Collaboration.}
You are a helpful medical assistant. \\
Your task is disease diagnosis. \\
You are given a \textbf{\{Modality\}} image. \\
The possible options are: \textbf{\{Label Set\}}. \\
The diagnoses of the most similar cases are \textbf{\{RAD\}}. \\
The reference answers by other models are \textbf{\{MoED\}}. \\
\\
You are a helpful medical assistant. \\
Your task is disease diagnosis. \\
You are required to diagnose the \textbf{\{Modality\}} image. \\
The reference diagnoses of the most similar cases are \textbf{\{RAD\}}. \\
The reference answers by other models are \textbf{\{MoED\}}. \\
The possible options are: \textbf{\{Label Set\}}. \\
\\
You are a helpful medical assistant. \\
You are required to classify the \textbf{\{Modality\}} Image. \\
The reference diagnoses of the most similar cases are \textbf{\{RAD\}}. \\
The reference answers by other models are \textbf{\{MoED\}}. \\
The available options are: \textbf{\{Label Set\}}. \\
\\
You are a helpful medical assistant. \\
You are required to diagnose the \textbf{\{Modality\}} image. \\
The reference diagnoses of the most similar cases are \textbf{\{RAD\}}. \\
The reference answers by other models are \textbf{\{MoED\}}. \\
The possible options are: \textbf{\{Label Set\}}. \\

\end{promptenv}
\caption{
Instructions for Generalist-Specialist Collaboration.
\textbf{\{Modality\}} is the placeholder for the name of different medical modalities.
\textbf{\{Label Set\}} denotes the candidate label set of the classification tasks.
\textbf{\{RAD\}} and \textbf{\{MoED\}} are the results of retrieval-augmented diagnosis and mixture-of-expert diagnosis.
}
\end{figure}

\begin{table}[h]
\centering
\caption{
The specifications of the ten selected computer vision models are detailed, encompassing the number of parameters, Giga Multiply-Add Operations per Second (GMACs), and the training time per epoch on a dataset, with PneumoniaMNIST serving as an example. 
Parameters are quantified in millions (M).
``IN1K'' refers to ImageNet-1K dataset~\cite{deng2009imagenet}.
}
\begin{tabular}{l c c c c }
    \toprule
    Model & Pretrained & Parameters (M) & GMACs & Training Time (sec/epoch)\\
    \midrule
    VGG16~\cite{simonyan2014very} & IN1K & 138.4  & 15.5 & 15.93\\ 
    AlexNet~\cite{krizhevsky2012imagenet} & IN1K & 62.3  & 0.1 & 3.34\\ 
    ResNet-18~\cite{he2016deep} & IN1K & 11.7  & 1.8 & 4.05 \\ 
    DenseNet-121~\cite{huang2017densely} & IN1K & 8.0  & 2.9 & 12.91\\ 
    EfficientNet-B4~\cite{tan2019efficientnet} & IN1K & 19.3  & 3.1 & 15.48\\ 
    ViT-B/16~\cite{dosovitskiy2020image} & IN1K &  86.6  & 16.9 & 19.35\\ 
    CLIP ViT-B/16~\cite{radford2021learning} & CLIP & 86.6  & 16.9 & 19.28\\ 
    EVA-02 ViT-B/16~\cite{fang2023eva} & CLIP & 86.3 & 16.9 & 23.74\\ 
    DINO ViT-B/16~\cite{caron2021emerging} & DINO & 85.8 & 16.9 & 19.14\\ 
    SAM ViT-B/16~\cite{kirillov2023segment} & SA-1B & 89.7 & 486.4 & 24.93\\ 
    \bottomrule
\end{tabular}
\label{tab:spemodel}
\end{table}

\begin{table}[h]
\caption{Hyperparameters used for training each specialist model. The specific parameters for each model include input dimension, hidden dimension, and dropout rate. The common parameters are batch size, number of epochs, optimizer, learning rate, scheduler, and weight decay.}
\label{tab:spehyper}
\centering
\begin{tabularx}{\textwidth}{l c c c c c c }
    \toprule
    & VGG16 & AlexNet & ResNet-18 & DenseNet-121 & EfficientNet-B4 & ViT-B/16 \\
    \midrule
    Input Dim & 224$\times$224 & 224$\times$224 & 224$\times$224 & 224$\times$224 & 224$\times$224 & 224$\times$224 \\
    Hidden Dim & 4096 & 4096 & 512 & 1024 & 1792 & 768 \\
    Dropout & 0.0 & 0.5 & 0.0 & 0.0 & 0.0 & 0.0 \\
    \bottomrule
\end{tabularx}

\vspace{1em} 

\begin{tabularx}{\textwidth}{l c c c c c }
    \toprule
    & CLIP ViT-B/16 & EVA-02 ViT-B/16 & DINO ViT-B/16 & SAM ViT-B/16 \\
    \midrule
    Input Dim & 224$\times$224 & 224$\times$224 & 224$\times$224 & 224$\times$224 \\
    Hidden Dim & 768 & 768 & 768 & 256 \\
    Dropout & 0.0 & 0.0 & 0.0 & 0.0 \\
    \bottomrule
\end{tabularx}

\vspace{1em} 

\begin{tabularx}{\textwidth}{X X X X X X X}
    \toprule
    Batch Size & Epochs & Optimizer & Learning Rate & Scheduler & Weight Decay & Cycle Limit \\
    \midrule
    64 & 100 & AdamW & 0.0001 & Cosine & 0.01 & 1\\
    \bottomrule
\end{tabularx}
\end{table}

\begin{table*}[!ht]
\centering
\caption{Data availability of training datasets. ``Open Access'' datasets are freely available to the public. For the ``Restricted Access'' datasets, please contact the respective dataset providers for access permissions. ``Credentialed Access'' datasets require specific permissions. ``Private'' datasets are not publicly accessible.}
\adjustbox{width=1\linewidth}{
\begin{tabular}{l|p{7.2cm}|l}
\toprule
\rowcolor{lightgray}
\textbf{Dataset Name} & \textbf{Link} & \textbf{Access} \\
\midrule
\rowcolor{lightgray!50}
\multicolumn{3}{c}{\textbf{Visual Question Answering} } \\
Slake-VQA~\cite{liu2021slake}&\url{https://www.med-vqa.com/slake/} &  Open Access \\
VQA-RAD~\cite{lau2018visual}&\url{https://osf.io/89kps/} &  Open Access \\
Path-VQA~\cite{he2020pathvqa}&\url{https://huggingface.co/datasets/flaviagiammarino/path-vqa} &  Open Access \\
PMC-VQA~\cite{zhang2023pmc}&\url{https://huggingface.co/datasets/xmcmic/PMC-VQA} &  Open Access \\
PMC-CaseReport~\cite{wu2023radfm}&\url{https://huggingface.co/datasets/chaoyi-wu/PMC-CaseReport} &  Open Access \\
\midrule
\rowcolor{lightgray!50}
\multicolumn{3}{c}{\textbf{Medical Report Generation} } \\
MIMIC-CXR~\cite{johnson2019mimiccxr}&\url{https://physionet.org/content/mimic-cxr/2.0.0/} & Credentialed Access  \\
IU-Xray~\cite{demner2016iu}&\url{https://www.kaggle.com/datasets/raddar/chest-xrays-indiana-university/data} &  Open Access \\
\midrule
\rowcolor{lightgray!50}
\multicolumn{3}{c}{\textbf{Medical Image Classification} } \\
VinDr-SpineXR~\cite{nguyen2021vindr} &\url{https://vindr.ai/datasets/spinexr} & Credentialed Access  \\
VinDr-PCXR~\cite{pham2022vindr} &\url{https://physionet.org/content/vindr-pcxr} & Credentialed Access  \\
VinDr-Mammo~\cite{nguyen2023vindr} &\url{https://vindr.ai/datasets/mammo} &  Credentialed Access \\
VinDr-CXR~\cite{nguyen2020vindrcxr} &\url{https://vindr.ai/datasets/cxr} & Credentialed Access  \\
CheXpert~\cite{irvin2019chexpert} &\url{https://stanfordmlgroup.github.io/competitions/chexpert/} & Restricted Access  \\
ChestX-ray14~\cite{wang2017chestx} &\url{https://nihcc.app.box.com/v/ChestXray-NIHCC} & Credentialed Access  \\
PCam200~\cite{kawai2023large} &\url{https://drive.google.com/drive/folders/1Oh7onawKsDW5ScamVO5ByXFgqdYJ39sK} & Open Access  \\
PAD-UFES-20~\cite{pacheco2020pad} &\url{https://data.mendeley.com/datasets/zr7vgbcyr2/1} &  Open Access \\
DermNet~\cite{kaggleDermnet} &\url{https://www.kaggle.com/datasets/shubhamgoel27/dermnet} &  Open Access \\
HAM10000~\cite{tschandl2018ham10000} &\url{https://www.kaggle.com/datasets/kmader/skin-cancer-mnist-ham10000} & Open Access  \\
ISIC2020~\cite{rotemberg2021patient} &\url{https://challenge2020.isic-archive.com/} & Open Access  \\
Kvasir~\cite{pogorelov2017kvasir} &\url{https://datasets.simula.no/kvasir/} &  Open Access \\
Kvasir Capsule~\cite{Smedsrud2021} &\url{https://datasets.simula.no/kvasir-capsule/} &  Open Access \\
WCE~\cite{kaggleWCE} &\url{https://www.kaggle.com/datasets/francismon/curated-colon-dataset-for-deep-learning} &  Open Access \\
GastroVision~\cite{jha2023gastrovision} &\url{https://github.com/DebeshJha/GastroVision} & Open Access  \\
ODIR~\cite{li2021benchmark} &\url{https://www.kaggle.com/datasets/andrewmvd/ocular-disease-recognition-odir5k} & Open Access  \\
Fundus1000~\cite{cen2021automatic} &\url{https://www.kaggle.com/datasets/linchundan/fundusimage1000} & Open Access  \\
RFMiD2.0~\cite{data8020029} &\url{https://zenodo.org/records/7505822} & Open Access  \\
Retinal OCT-C8~\cite{9740985} &\url{https://www.kaggle.com/datasets/obulisainaren/retinal-oct-c8} & Open Access  \\
UltraBreast & - & Private\\
\hline
\end{tabular}
}
\label{data_available_training}
\vspace{60pt}
\end{table*}

\begin{table*}[!ht]
\centering
\caption{Data availability of benchmark datasets.
``Open Access'' datasets are freely available to the public.
``Credentialed Access'' datasets require specific permissions.}
\adjustbox{width=1\linewidth}{
\begin{tabular}{l|p{7cm}|l}
\toprule
\rowcolor{lightgray}
\textbf{Dataset Name} & \textbf{Link} & \textbf{Access} \\
\midrule
\rowcolor{lightgray!50}
\multicolumn{3}{c}{\textbf{Visual Question Answering} } \\
Slake-VQA~\cite{liu2021slake}&\url{https://www.med-vqa.com/slake/} &  Open Access \\
VQA-RAD~\cite{lau2018visual}&\url{https://osf.io/89kps/} &  Open Access \\
Path-VQA~\cite{he2020pathvqa}&\url{https://huggingface.co/datasets/flaviagiammarino/path-vqa} &  Open Access \\
PMC-VQA~\cite{zhang2023pmc}&\url{https://huggingface.co/datasets/xmcmic/PMC-VQA} &  Open Access \\
VQA-Med~\cite{ImageCLEFVQA-Med2019}&\url{https://github.com/abachaa/VQA-Med-2019} &  Open Access \\
OmniMedVQA~\cite{hu2024omnimedvqa}&\url{https://openxlab.org.cn/datasets/GMAI/OmniMedVQA} &  Credentialed Access \\
\midrule
\rowcolor{lightgray!50}
\multicolumn{3}{c}{\textbf{Medical Report Generation} } \\
MIMIC-CXR~\cite{johnson2019mimiccxr}&\url{https://physionet.org/content/mimic-cxr/2.0.0/} & Credentialed Access  \\
IU-Xray~\cite{demner2016iu}&\url{https://www.kaggle.com/datasets/raddar/chest-xrays-indiana-university/data} &  Open Access \\
\midrule
\rowcolor{lightgray!50}
\multicolumn{3}{c}{\textbf{Medical Image Classification} } \\
PCam200~\cite{kawai2023large} &\url{https://drive.google.com/drive/folders/1Oh7onawKsDW5ScamVO5ByXFgqdYJ39sK} & Open Access  \\
Dermnet~\cite{kaggleDermnet} &\url{https://www.kaggle.com/datasets/shubhamgoel27/dermnet} &  Open Access \\
HAM10000~\cite{tschandl2018ham10000} &\url{https://www.kaggle.com/datasets/kmader/skin-cancer-mnist-ham10000} & Open Access  \\
RetOCT~\cite{9740985} &\url{https://www.kaggle.com/datasets/obulisainaren/retinal-oct-c8} & Open Access  \\
VinDr-SpineXR~\cite{nguyen2021vindr} &\url{https://vindr.ai/datasets/spinexr} & Credentialed Access  \\
VinDr-PCXR~\cite{pham2022vindr} &\url{https://physionet.org/content/vindr-pcxr/1.0.0/} & Credentialed Access  \\
VinDr-Mammo~\cite{nguyen2023vindr} &\url{https://vindr.ai/datasets/mammo} &  Credentialed Access \\
ChestMNIST~\cite{yang2023medmnist} &\url{https://medmnist.com/} & Open Access \\

PneumoniaMNIST~\cite{yang2023medmnist} &\url{https://medmnist.com/} & Open Access \\
BreastMNIST~\cite{yang2023medmnist} &\url{https://medmnist.com/} & Open Access \\
OrganAMNIST~\cite{yang2023medmnist} &\url{https://medmnist.com/} & Open Access \\
PathMNIST~\cite{yang2023medmnist} &\url{https://medmnist.com/} & Open Access \\
OCTMNIST~\cite{yang2023medmnist} &\url{https://medmnist.com/} & Open Access \\
CBIS-DDSM(MASS)~\cite{sawyer2016curated}&\url{https://www.kaggle.com/datasets/awsaf49/cbis-ddsm-breast-cancer-image-dataset} & Open Access\\
CBIS-DDSM(CALC)~\cite{sawyer2016curated}&\url{https://www.kaggle.com/datasets/awsaf49/cbis-ddsm-breast-cancer-image-dataset} & Open Access\\
FMC-Colon~\cite{wang2023real}&\url{https://github.com/openmedlab/MedFM} & Credentialed Access\\
FMC-Endo~\cite{wang2023real}&\url{https://github.com/openmedlab/MedFM} & Credentialed Access\\
FMC-Chest~\cite{wang2023real}&\url{https://github.com/openmedlab/MedFM} & Credentialed Accesss\\
Derm7pt~\cite{kawahara2018seven}&\url{https://derm.cs.sfu.ca/Welcome.html} & Credentialed Accesss\\
BRSET~\cite{nakayama2023brazilian}&\url{https://physionet.org/content/brazilian-ophthalmological/1.0.0/} & Credentialed Accesss\\
\bottomrule
\end{tabular}
}
\label{data_available_testing}
\end{table*}
\begin{table*}
\centering
\caption{The public code used in this study.}
\adjustbox{width=1\linewidth}{
\begin{tabular}{l|p{10cm}}
\toprule
\textbf{Name} & \textbf{URL} \\
\midrule
InternVL~\cite{chen2023internvl}&\href{https://github.com/OpenGVLab/InternVL}{https://github.com/OpenGVLab/InternVL}\\
RadFM~\cite{wu2023radfm}&\href{https://github.com/chaoyi-wu/RadFM}{https://github.com/chaoyi-wu/RadFM}\\
LLaVA-Med~\cite{li2024llavamed}&\href{https://github.com/microsoft/LLaVA-Med}{https://github.com/microsoft/LLaVA-Med}\\
Med-Flamingo~\cite{moor2023medflamingo}&\href{https://github.com/snap-stanford/med-flamingo}{https://github.com/snap-stanford/med-flamingo}\\
MedMNIST~\cite{yang2023medmnist}&\href{https://github.com/MedMNIST/MedMNIST}{https://github.com/MedMNIST/MedMNIST}\\
MedMNIST+~\cite{doerrich2024rethinking}&\href{https://github.com/sdoerrich97/rethinking-model-prototyping-MedMNISTPlus}{https://github.com/sdoerrich97/rethinking-model-prototyping-MedMNISTPlus}\\
MultiMedEval~\cite{royer2024multimedeval}&\href{https://github.com/corentin-ryr/MultiMedEval}{https://github.com/corentin-ryr/MultiMedEval}\\
\bottomrule
\end{tabular}
}
\label{tab:public_code} 
\end{table*}
\end{appendices}



\end{document}